\documentclass[10pt]{article}
\usepackage{float}
\usepackage[english]{babel}
\usepackage{subcaption}
\usepackage{tikz-cd,tikz}
\usepackage{tikz-qtree}
\usepackage[letterpaper,top=2cm,bottom=2cm,left=2cm,right=3cm,marginparwidth=1.75cm]{geometry}
\usepackage{booktabs} 
\usepackage{algorithm}
\usepackage{algorithmic}
\usepackage{amsmath}
\usepackage{amssymb}
\usepackage{graphicx}
\usepackage{multicol,multirow}
\usepackage[colorlinks=true, allcolors=blue]{hyperref}
\usepackage{color,xcolor}
\usepackage{float}
\usepackage{cite}
\usepackage{url}

\usepackage{pifont}%
\newcommand{\cmark}{\ding{51}}%
\newcommand{\xmark}{\ding{55}}%

\title{LeMON: \textbf{Le}arning to Learn \textbf{M}ulti-\textbf{O}perator \textbf{N}etworks}
\author{Jingmin Sun \thanks{Department of Mathematics, Carnegie Mellon University, 5000 Forbes Avenue Pittsburgh, PA 15213. \texttt{Email: jingmins@andrew.cmu.edu.}}
	\and
	Zecheng Zhang \thanks{Department of Mathematics, Florida State University, 222 S Copeland St, Tallahassee, FL 32306.  \texttt{Email: zecheng.zhang.math@gmail.com.}}
	\and
	Hayden Schaeffer \thanks{Department of Mathematics, UCLA, 520 Portola Plaza Math Sciences Building 6363, Los Angeles, CA 90095. \texttt{Email: hayden@math.ucla.edu.}  }
}
\date{}

\begin{document}
	\maketitle
	\begin{abstract}
		Single-operator learning focuses on training a neural network to approximate a specific operator, while multi-operator learning extends this capability by training a single model to approximate multiple distinct operators simultaneously. This broader scope enables better generalization to out-of-distribution tasks and supports predictions across a wider range of physical phenomena.
		In this work, we introduce LeMON, a meta-learning framework that transforms a single-operator model into a multi-operator learner and enhances the multi-operator learner with the improved capability of solving challenging, unseen PDE-related problems. 
		Two key insights drive the success of LeMON's fast adaptation: (1) leveraging diverse PDE families during pre-training to improve generalization (data diversity) and (2) employing meta-learning to optimize training and fine-tuning strategies (learning to learn).
		Furthermore, by incorporating a symbolic operator encoding module into the neural operator network, LeMON further improves predictive accuracy through symbolically induced, learnable learning rates.
		Numerical experiments demonstrate that LeMON is effective across a variety of neural operator architectures—including Deep Operator Networks, Fourier Neural Operators, and more general Transformer-based models—consistently outperforming standard training and fine-tuning methods.
		Significant performance gains are also observed when symbolic meta-learning components are applied to these state-of-the-art architectures to learn adaptive learning rate (LeMONS).
		
		\let\thefootnote\relax\footnotetext{The code is available at: \url{https://github.com/JingminSun/LeMON}.}
		
	\end{abstract}
	
	\section{Introduction}
	Developing robust numerical solvers for partial differential equations (PDEs) is a crucial topic in scientific and engineering applications. Scientific Machine Learning (SciML) methods have emerged as effective options for solving PDEs under a variety of challenging conditions. Compared to standard numerical algorithms, SciML methods \cite{raissi2019physics,karniadakis2021physics, lu2021learning, zhang2023belnet, sun2020neupde, schaeffer2017learning, jin2022mionet, li2022fourier, lin2023b, schaeffer2013sparse, schaeffer2018extracting, schaeffer2017sparse, efendiev2022efficient} are efficient during the inference stage and can be utilized in scarce or limited data regimes.
	When the target equation lies within the training distribution, machine learning methods can provide reliable and fast predictions. SciML methods also excel in simulating and solving real engineering applications when both the equation of the physical system (such as PDEs or ODEs) and observed data are available.
	
	SciML methods can be categorized into two types. 
	The first type includes data-free solvers, such as Physics-Informed Neural Networks (PINNs) \cite{raissi2019physics,karniadakis2021physics, leung2022nh, zhu2019physics}. 
	These methods do not require additional data and thus approximate the PDE solutions using parameterized neural networks. 
	However, these methods solve one equation at a time, meaning a new network must be trained for each new equation, which does not fully leverage the benefits of SciML methods. Single operator learning (SOL) methods \cite{lu2021learning, lu2022comprehensive, lin2023b, lu2022multifidelity, li2019diffusion, li2020fourier, li2020neural, zhang2023belnet, zhang2024d2no, zhang2023discretization, moya2024conformalized,yin2024dimon} approximate the solution operator $G: U\rightarrow V$ that maps between two function spaces $U, V$, making them ideal candidates for solving PDEs, as solving a PDE can be formulated as evaluating the solution operator. 
	For example, when solving time-dependent PDEs given initial conditions (ICs), one can first train a deep neural operator (DNO) using pre-generated data consisting of input functions (ICs) and output functions (solutions at later times). 
	During the inference stage, to solve the same PDE with a new IC from a sample related to the training input function distribution, one can simply input the new IC into the trained deep neural operator.
	Another typical example of leveraging DNO to solve PDE is using limited fine-scale high-fidelity data to enhance a low-accuracy solution \cite{zhang2023bayesian, lu2022multifidelity, howard2022multifidelity}. 
	In this setting, the input functions are coarse-scale solutions, which are easy to obtain, while the output functions are a limited number of fine-scale (high-fidelity) observations from scarce sensors. The DNO learns the downscaling mapping in order to significantly enhances the coarse-scale models.

	Many DNOs have been proposed and have shown success in solving various scientific and engineering problems, such as those in geology, physics, and climate science \cite{pathak2022fourcastnet, zhu2023fourier, jiang2023fourier, li2023solving, lin2023b, moya2023approximating, chen2023neural}. 
	However, challenges remain when applying DNOs to new (unseen) tasks. We define a new task as one where the new PDE or operator differs from the operators that are in the training set, or the input function distributions differ from the training distributions.
	This means that the retraining is again required to solve new PDEs, which limits the utility of DNO for solving PDEs.
	In \cite{zhang2024deeponet}, the authors introduced a fine-tuning approach that adapts a pre-trained model using a physics-informed loss, enabling it to address previously unseen problems. While effective in certain cases, this method exhibits limited generalization when the target task differs substantially from the training distribution or the PDEs present high-order derivatives.
	In this work, our first contribution is the development of a meta-learning framework that enables DNOs, including the popular Deep Operator Network (DeepONet), Fourier Neural Operators (FNO), transformed-based model, etc., to efficiently adapt to new tasks. The meta-learning algorithms will be reviewed in detail later.

	One approach for applying DNOs to new tasks is through fine-tuning, which can be effective for similar tasks but requires additional training and data. In \cite{zhang2024deeponet}, it was shown that fine-tuning can be made data-free using a PINNs formulation. Recently multi-operator learning (MOL) \cite{sun2025towards, liu2024prose, mccabe2023multiple, yang2023context, yang2023prompting, zhang2024modno} approaches have been proposed; specifically, in \cite{sun2025towards, liu2024prose}, the authors show that MOL is able to accurately handle new tasks.
	MOL involves constructing a single neural network architecture to learn multiple operators $G_1, G_2, ..., G_{N_{op}}$ simultaneously. 
	Specifically, MOL is trained using data, comprising of input function and output function pairs, from multiple different operators. MOL can be categorized into two classes: (1) non-operator encoding and (2) operator encoding.
	The non-operator encoding approaches \cite{zhang2024modno, rahman2024pretraining, subramanian2024towards} utilizes information from the input functions of different operators, and can only handle a limited number of families of PDEs, which are generated by varying free parameters. Without an operator encoding strategy, these approaches may not be able to predict operators which were not seen in the training dataset. In contrast, neural networks that use an operator encoding approach \cite{liu2024prose, yang2023prompting, sun2025towards} incorporate an embedding of the operator (the equation, a label, etc.) alongside the corresponding input functions, which usually results in a better generalization. These approaches are also steps towards a PDE foundation model. In particular, the additional encoding allows the model to handle new PDE tasks in a zero-shot manner, as demonstrated by \cite{sun2025towards}.
	For example, the PROSE model \cite{sun2025towards, liu2024prose} shows that a neural network can generalize to new PDEs not seen during training by leveraging multi-modality inputs and outputs. In the PROSE generalization tests, the new PDEs share similar terms to those in the training set, which is expected for zero-shot training. 
	
	MOL has demonstrated the potential to address out-of-distribution (OOD) tasks without the need for retraining. 
	However, the underlying mechanism that enables such generalization remains not well understood. In this work, our second contribution is demonstrating that data diversity is a key factor for generalization. 
	Specifically, we show that training MOL models on data derived from a wide range of operators exhibiting distinct properties significantly enhances their ability to generalize. This further shows the significance of utilizing MOL model to solve scientific computing related problems. Notably, we demonstrate that a model pre-trained with diverse PDE data can be quickly fine-tuned to solve downstream tasks using only limited samples, i.e., an amount of data that would be insufficient for the training of a DNO model from scratch.
	
	While data diversity is crucial, the pre-training and fine-tuning strategy used to obtain a good initialization and adaptation to new problems is equally important. 
	Ideally, the pre-training process should identify network parameters that are most sensitive to task variations, as emphasized in \cite{finn2017model}. These sensitive parameters should be prioritized during fine-tuning, enabling rapid adaptation to new tasks. Meanwhile, parameters encoding shared features across different tasks should remain largely unchanged, preserving the model’s generalization ability.
	To optimize network initialization for adapting to new PDE operators, we propose a meta-learning algorithm \cite{finn2017model, sun2019meta, antoniou2019train, santoro2016meta, ren2018meta, snell2017prototypical, vinyals2016matching, oreshkin2018tadam} that trains a base learner capable of efficient adaptation using minimal downstream data. This forms our third contribution: the design of a learning-to-learn framework for pre-training MOL models.
	
	Moreover, our approach enables a single-operator DNO, originally designed to approximate one operator at a time, to handle multiple operator tasks via few-shot adaptation. 
	Specifically, our meta-learning algorithm identifies the parameters most relevant for adaptation by training the DNO on data from a variety of operators. During adaptation, the model fine-tunes in a manner that mirrors the meta-training process, allowing for rapid convergence on new tasks with limited data.
	Single-operator DNOs are inherently designed to approximate a single operator at a time, which limits their ability to leverage data diversity and thus hinders generalization to new tasks. Our proposed algorithm does not violate the theoretical foundations of single-operator networks. Instead, it provides a well-initialized model that can be efficiently adapted to previously unseen operator tasks through tailored training and adaptation strategies.
	Notably, we validate the effectiveness of our algorithm across several widely used DNO architectures, including DeepONet and FNO, as well as recently developed transformer-based DNOs featured in state-of-the-art MOL frameworks such as PROSE \cite{liu2024prose, sun2025towards} or BCAT \cite{liu2025bcat}.
	
	Additional, the learning rate plays a critical role in learning to learn. To further improve adaptation performance, we introduce a learnable and \textit{adaptive learning rate strategy}, which is achieved through a symbolic encoding of the operator. This adaptive scheme allows the learning rate to dynamically adjust during the training process based on the properties of the new operator task.
	The result is an effective use of the the symbolic operator encoding framework, originally introduced in \cite{sun2025towards, liu2024prose}, in capturing essential structural information of operators, and thus facilitating extrapolation to previously unseen tasks.
	Since our algorithm enables the solution of new problems through specific training and fine-tuning with curated datasets, we refer to it as \textit{Learning to Learn Multi-Operator Networks (LeMON)}.
	We summarize the contribution of the work as follows.

	\begin{enumerate}
		\item We show that data diversity is a key mechanism driving the generalization capabilities of MOL. This finding holds for both MOL models with operator encoding structures (e.g., PROSE) and those without, including other transformer-based MOL architectures \cite{yang2023prompting, mccabe2023multiple}.
		Since MOL can handle data from operators with different properties, we show the potential of leveraging a pre-trained MOL model as a base learner for addressing new PDE problems through the proposed learning-to-learn methods in the few-shot setting.
		\item We design a meta-learning-based algorithm that enables single operator learning DNOs, originally designed to only approximate a single PDE operator, to be trained with data from multiple PDE operators. 
		This approach enables the model to identify parameters that are most sensitive to variations in the data, allowing for efficient fine-tuning that emulates the meta-training process. As a result, the model can quickly attain improved accuracy even when only limited data is available for fine-tuning. The proposed method also enhances the pre-training and adaptation performance of MOL models when compared to conventional network training approaches.
		\item We propose a learnable inner-loop learning rate strategy, implemented via a symbolic operator encoding module, to enhance our meta-learning framework. This strategy leads to significant improvements in prediction accuracy. The proposed algorithm integrates the symbolic module into various neural operator architectures and consistently enhances their generalization performance. 
	\end{enumerate}

	The rest of the paper is organized as follows.
	In Section \ref{sec_overview}, we review the SOL and MOL and related-ML works for solving PDEs.
	In Section \ref{sec_methods}, we introduce the methodology for LeMON and LeMON with symbolic adaption (LeMONS). To validation the proposed methods, we conduct numerical experiments and present the results in Section \ref{sec_exp}. Some future research directions are discussed in Section \ref{sec_conclusion}.
	
	\section{Related Works and Overview}\label{sec_overview}
	Learning solution operators for partial differential equations via machine learning has emerged as a paradigm for accelerating simulations. Given a parametric PDE where inputs $ \{u_0(x), q\} $ (e.g., initial conditions, coefficients) map to solutions $ u(x,t) $ via an operator $ {G} $, operator learning aims to approximate $ {G} $ using neural networks $ {G}_{\theta} $ trained on input-output pairs. 
	
	\subsection{Single Operator Learning}\label{subsec:sol}
	The analysis and underlying structure of operator learning is supported by approximation theory. A landmark result by~\cite{chen1995universal} established that neural networks can serve as universal approximators of nonlinear operators, not just finite-dimensional functions. DeepONet~\cite{lu2019deeponet}, one of the earliest framework for operator learning, employs dual subnetworks: a branch network encodes input functions into coefficients, while a trunk network evaluates basis functions at query locations, combining them linearly to predict solutions via 
	\[\label{eq:deeponet}
	{{G}}_\theta(u_0)(y) = \sum_{k=1}^p b_k(u_0;\theta_{\text{br}}) t_k(y;\theta_{\text{tr}}).
	\]where $u_0(\cdot)$ denotes the input of the operators (i.e. the initial condition of the PDE), and $y = (x,t)$ denotes the query point of our operator outputs, and  $ \theta = (\theta_{\text{br}}, \theta_{\text{tr}}) $ denotes trainable parameters from branch net and trunk net respectively. Fourier Neural Operators \cite{li2020fourier} utilizes the Fourier transform to learn global convolution kernels in frequency space, excelling in structured domains. However, using the standard training process, these frameworks face some limitations including an inability to learn distinct operators using the same network and the lack of generalization to new PDE tasks outside of those seen in training.
	
	
	\subsection{Multi-operator learning} \label{subsec:prose}
	
	Multi-Operator Learning extends the operator learning paradigm to handle multiple PDEs or physical systems within one unified model. Formally, suppose we have $N_{\text{op}}$ different operators ${G}_1,\dots,{G}_{N_{\text{op}}}$ such that for each $i$, a unified MOL network $\mathcal{M}$ can approximate ${G}_i$. In practice, we provide the training procedure with data pooled from all operators $\{u, {G}_k(u)\}$ for $k=1,\dots,N_{\text{op}}$, and minimize a total loss over all operators. While a direct approach may be to add an identifier (i.e., label) to the input $k$ to indicate which operator the network should utilize, such processes lead to models that transfer to unseen systems. A major insight from recent studies is that operator encoding is crucial for successful MOL. If we simply mix data from multiple operators without signaling to the model which operator each data point comes from, the network often interpolates a fit to an ``average” operator that does not correspond to any particular system. As PROSE (Predicting Operators and Symbolic Expressions) \cite{liu2024prose,sun2025towards} note, without an encoding, a multi-operator network may be unable to predict operators not seen in the training set. In contrast, adding an explicit operator embedding, i.e., some representation of the PDE or system as part of the input, greatly improves the model’s ability to distinguish and generalize.

	PROSE achieve the MOL tasks by leveraging joint operator learning and model discovery. Its transformer architecture processes bimodal inputs: (1) operator inputs and (2) symbolic PDE descriptions. PROSE outputs both the solution and a symbolic form of the PDE, enabling zero-shot generalization to unseen equations by incorporating their symbolic structure during inference. PROSE demonstrates that encoding PDE structure into operator learning frameworks enhances generalization, reduces data dependence, and supports interpretability. In this work, we will only discuss the solution operator prediction task, and refer to \cite{liu2024prose,sun2025towards} for more details on symbolic discovery.

	\subsection{Meta-Learning}\label{subsec:meta}
	
	Meta-learning, often referred to as ``learning to learn," involves systems that adapt their learning mechanisms by leveraging knowledge gained from prior experiences or other tasks \cite{lemke2015metalearning}. Traditional supervised learning focuses on training models for specific tasks using large, fixed datasets. In meta-learning, the goal is toward handling a variety of interrelated tasks, each typically accompanied by limited task-specific data. The main goal is to distill higher-level knowledge from the learning processes across these tasks, enhancing the system's ability to rapidly adapt to new, unseen challenges with minimal data.
	
	Current meta-learning approaches are broadly categorized into three categories: (1) Model-based methods, which employ cyclic or memory-augmented neural networks (e.g., LSTMs) to dynamically adapt their states using task-specific training sequences, as seen in memory-augmented networks~\cite{santoro2016meta} and neural attentive meta-learners~\cite{mishra2017simple}; (2) Metric-based techniques, which leverage learned embeddings and non-parametric algorithms (e.g., nearest-neighbor classification) to define task-adaptive similarity metrics, for example, prototypical\cite{snell2017prototypical}, matching\cite{vinyals2016matching}, and relation networks\cite{sung2018learning}; and (3) Optimization-based strategies, such as model-agnostic meta-learning (MAML)\cite{finn2017model}, which infer shared optimization dynamics across tasks to enable rapid adaptation through optimized parameter initializations and hyperparameters. Specifically, the authors of MAML \cite{finn2017model} proposed an optimization algorithm that provides an optimal initialization for model parameters, enabling fast adaptation to new tasks. It operates in two phases:
	\begin{itemize}
		\item \textbf{Inner Loop}: Task-specific adaptation using few gradient steps on support data,
		\item \textbf{Outer Loop}: Meta-update optimizing cross-task performance after adaptation.
	\end{itemize}
	We extend MAML to PDE operator learning by defining tasks as distinct PDE solution operators  ${G}:U\to V $, incorporating symbolic encodings to enhance adaptation.

	\section{Methodology}\label{sec_methods}
	\subsection{Problem Formulation}\label{subsec:formulation}
	Consider the PDE family:
	\begin{align}\label{eq:pde}
	\mathcal{G}(u; q) &= 0, \quad x \in \Omega, \ t \in [0,T], \ q \sim \mathcal{D}_q \\
	u(x,0) &= \mathcal{I}(x; s), \quad s \sim \mathcal{D}_s
	\end{align}
	with boundary conditions. Here $ \mathcal{G} $ represents a parameterized PDE family with $ q \in \mathcal{D}_q $, and $ \mathcal{I}(x;s)\in I $ generates initial conditions parameterized by $ s \in \mathcal{D}_s $. For example, Burgers' equation $ \mathcal{G}(u;q) = u_t - q(u^2/2)_x = 0 $ forms a PDE family through $ q $-variation. The goal is to approximate the solution operator $ G: I \rightarrow {V} $ mapping initial conditions to solutions.
	
	\subsection{Multi-Operator Learning Framework}\label{subsec:mol}
	Standard SOL methods (e.g., DeepONet or FNO) train on data $ \{\mathcal{I}(x;s)\} $ and a fixed PDE  with $ \mathcal{G}_1(u;q_1) $, limiting applicability to new operators with $q_2\neq q_1$ and new PDEs $ \mathcal{G}_0 \neq \mathcal{G}_1 $. 
	A MOL framework proposes to solve this issue with an additional input via an operator encoding $ \mathcal{E} $~\cite{sun2025towards}: For PDE solution operators $ \{G_i: {I}_i \rightarrow {V}_i\}_{i=1}^{N_{op}}$, MOL approximates $ G_i(u_0) $ via
	\begin{equation}
	{G}_{i,\theta}(u_0) = \mathcal{M}(u_0, \mathcal{E}(G_i))
	\end{equation}
	where $u_0(\cdot)\equiv u(\cdot,0)= \mathcal{I}_i(\cdot,s)\in {I}_i$ and $ \mathcal{M} $ is an MOL model whose inputs are both the initial data and an operator encoding scheme.
	
	\subsection{LeMON: Learning to learn multi-operator networks }
	\label{subsec:lemon}
	In this section, we introduce our proposed training and tuning algorithm, LeMON, which enables the training of a single-operator learning network using data from multiple operators. 
	This approach significantly enhances the network’s adaptation capability, measured by the post-tuning prediction relative error when only limited downstream data from a new task is available.
	
	In practical settings, computational constraints and limited access to symbolic PDE representations could make direct operator encoding difficult. To address this problem, we propose solving the MOL problem using SOL architectures with a meta-learning training algorithm.
	The network is trained to identify parameters that are sensitive to changes in the data (i.e., PDE operators). During fine-tuning, we mirror the inner-loop updates used in training (Lines 4 to 8 of Algorithm~\ref{algo:LeMON1}) but apply them to the new task data. This allows the network to adapt by updating the parameters that are most responsive to the specific characteristics of task~$i$, consistent with the training procedure.
	
\begin{figure}[H]
		\vskip 0.2in
		\begin{center}
			\small
			\centerline{\includegraphics[width=.8\linewidth]{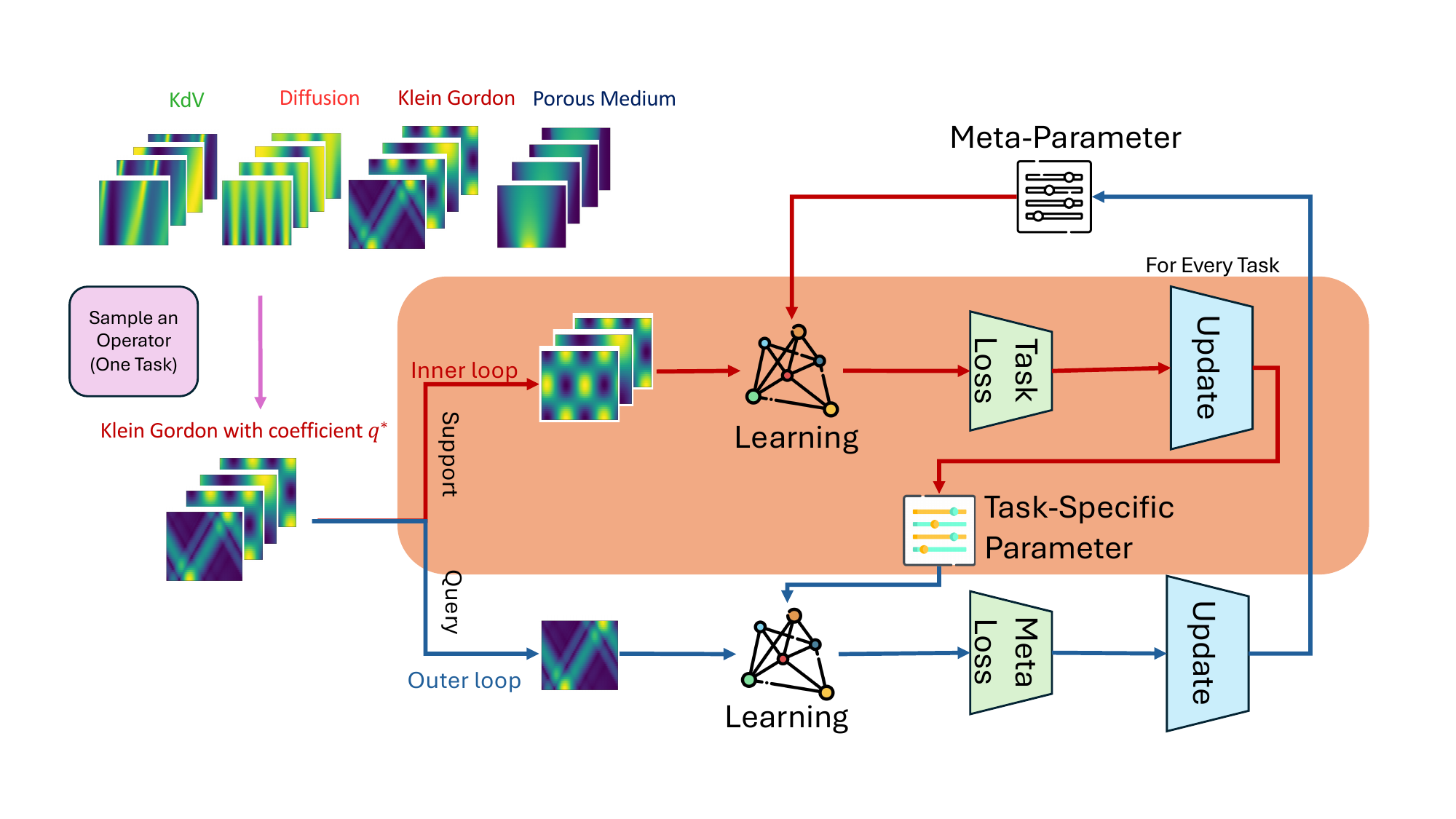}}
			\caption{\textbf{LeMON algorithm illustration}. Inner loop: Task (Operator)- specific updates; Outer loop: Meta updates}
			\label{fig:lemonillustration}
		\end{center}
		\vskip -0.2in
	\end{figure}

	We extend model-agnostic meta-learning (MAML) to  SOL architectures, enabling rapid adaptation to new PDE tasks. At each meta-training iteration, a batch of data associated to $ N $ PDEs $ \{\mathcal{G}_i(u; q_i)\}_{i=1}^N$ is sampled. For each PDE task $ \mathcal{G}_i $, the model undergoes $ p $ inner-loop adaptations using a support dataset $ D_i $, where task-specific parameters $ \theta_{i,j} $ are updated from the meta parameter $\theta$ via gradient descent on the loss $ \mathcal{L}_{\mathcal{G}_i}^{\mathcal{M}}(D_i; \theta_{i,j}, \theta) $, (with $\theta_{i,0}= \theta$), defined as the mean squared error (MSE) between the model’s predictions and ground-truth solutions:  
	\begin{equation}
	\mathcal{L}_{\mathcal{G}_i}^{\mathcal{M}}(D; \theta_{\text{adapted}}, \theta) = \mathbb{E}_{(u, v) \sim D} \left[ \left\| \mathcal{M}(u, \mathcal{E}(G_i); \theta_{\text{adapted}}) - v \right\|^2 \right].
	\end{equation}
	After adaptation, a meta-update optimizes the global meta parameters $ \theta $ by evaluating the adapted models on query datasets $ D_i' $, backpropagating through the $ p $-step adaptation process. This two-level optimization encourages $ \theta $ to learn a parameter initialization that generalizes across operators, allowing the SOL architecture to applied to unseen PDEs with minimal task-specific data. By meta-learning over diverse operators, LeMON circumvents the need for explicit symbolic PDE representations while maintaining compatibility with standard SOL frameworks like DeepONet or FNO. We detailed the algorithm in Algorithm~\ref{algo:LeMON1} and illustrated it in Figure~\ref{fig:lemonillustration}.

	\begin{algorithm}[H]
		\caption{Data Sampling}
		\label{algo:datasampling}
		\textbf{Input}: PDE family $\mathcal{G}(u; q)$ , empty dataset $D$, $N_q$, and $N_u$\\
		\textbf{Output}: Dataset $D$
		
		\begin{algorithmic}[1]
			\STATE Sample $N_q$ data associated to PDE $\mathcal{G}(u; q)$ .
			\STATE Sample $N_u$ initial conditions $u$ by sampling $s \sim \mathcal{D}_s$
			\STATE Append all input $u$ and output $G(u)$ into dataset $D$
			\STATE \textbf{return} $D$
		\end{algorithmic}
	\end{algorithm}

	\begin{algorithm}[H]
		\caption{LeMON training}
		\label{algo:LeMON1}
		\textbf{Input}: Operator batch size $N$, inner loop step $p$, a base model $\mathcal{M}$\\
		\textbf{Parameters}: Randomly initialized network parameters $\theta$, inner loop adaptation rate $\eta$, and meta update learning rate $\eta_m$\\
		\textbf{Output}: Updated network parameters $\theta$
		
		\begin{algorithmic}[1]
			\FOR{$k = 1$ {\bfseries to} End}
			\STATE Sample a batch of data associated to $N$ PDEs $\{\mathcal{G}_i(u; q_i)\}_{i=1}^N$
			\FOR{$i = 1,2,\cdots, N$}
			\STATE Obtain support dataset $D_i$ by Algorithm~\ref{algo:datasampling} associated with  $\mathcal{G}_i(u; q_i)$ for local update
			\STATE Set $\theta_{i,0} = \theta$
			\FOR{$j = 1,2,\cdots, p$}
			\STATE $\Delta_j = \nabla_{\theta_{i,j-1}}\mathcal{L}_{\mathcal{G}_i}^{\mathcal{M}}(D_i; \theta_{i,j-1};\theta)$ using $D_i$ 
			\STATE$\theta_{i,j} = \theta_{i,j-1} - \eta \Delta_j$
			\ENDFOR
			\STATE Obtain query dataset $D_i'$ by Algorithm~\ref{algo:datasampling} associated with  $\mathcal{G}_i(u; q_i)$ for meta update
			\ENDFOR
			\STATE $\theta \leftarrow \theta - \eta_m \sum_{i} \nabla_{\theta} \mathcal{L}_{\mathcal{G}_i}^{\mathcal{M}}(\mathcal{D}_i';\theta_{i,p};\theta)$
			\ENDFOR
		\end{algorithmic}
	\end{algorithm}
	Notably, in Algorithm~\ref{algo:LeMON1}, we refer to Lines 7-8  as the inner-loop updates, and Line 12 as the outer-loop update (i.e., the meta-update). The inner-loop update aims to adjust the parameters specific to a given task~$i$, or equivalently, to a specific PDE operator~$\mathcal{G}_i$. For each task~$i$, the inner-loop update begins from the previously saved meta-parameters~$\theta_{i,0}=\theta$, and updates the inner-loop parameters~$\theta_{i,j}$ by support dataset $D_i$. In contrast, the meta-update computes the gradient and updates the initial parameters~$\theta$ using an aggregation of the gradients from most-recent parameter $\theta_{i,p}$ with the query dataset~$\mathcal{D}_i^{\prime}$. The detailed calculation is presented in Eq.~\eqref{chain_1}, and we use the first order approximation in Eq,~\eqref{foapprox} \cite{finn2017model}:
	\begin{align}
	\nabla_{\theta} \mathcal{L}_{\mathcal{G}_i}^\mathcal{M}(\mathcal{D}_i';\theta_{i,p};\theta) &=    \nabla_{\theta_{i,p}} \mathcal{L}_{\mathcal{G}_i}^\mathcal{M}(\mathcal{D}_i';\theta_{i,p};\theta)\nabla_{\theta}\theta_{i,p}\label{chain_1} \\
	\nabla_{\theta}\theta_{i,p}&= \nabla_{\theta}\theta_{i,p-1} - \eta\nabla_{\theta}\Delta_j
	\\
	&\approx  \nabla_{\theta}\theta_{i,p-1}\approx \cdots\approx I_{\text{dim}_\theta}  \label{foapprox}
	\end{align}
	
	\subsection{Symbolic-Aware Adaptation Rate}\label{subsec:symbolic_lr}

	A key aspect to address in the training of MOL systems is that using a range of PDE problems and model architectures often necessitate distinct inner-loop adaptation learning rates $ \eta $ to maximize adaptation efficiency. While the vanilla LeMON employs a fixed $ \eta $ across tasks, this uniform approach may be suboptimal for heterogeneous operators. 
	To address this, we propose a learnable learning rate strategy by leveraging symbolic encodings of PDE operators (e.g., equation structures or coefficients), which provide task-specific prior knowledge.

\begin{figure}[H]
		\vskip 0.2in
		\begin{center}
			\centerline{\includegraphics[width=\linewidth]{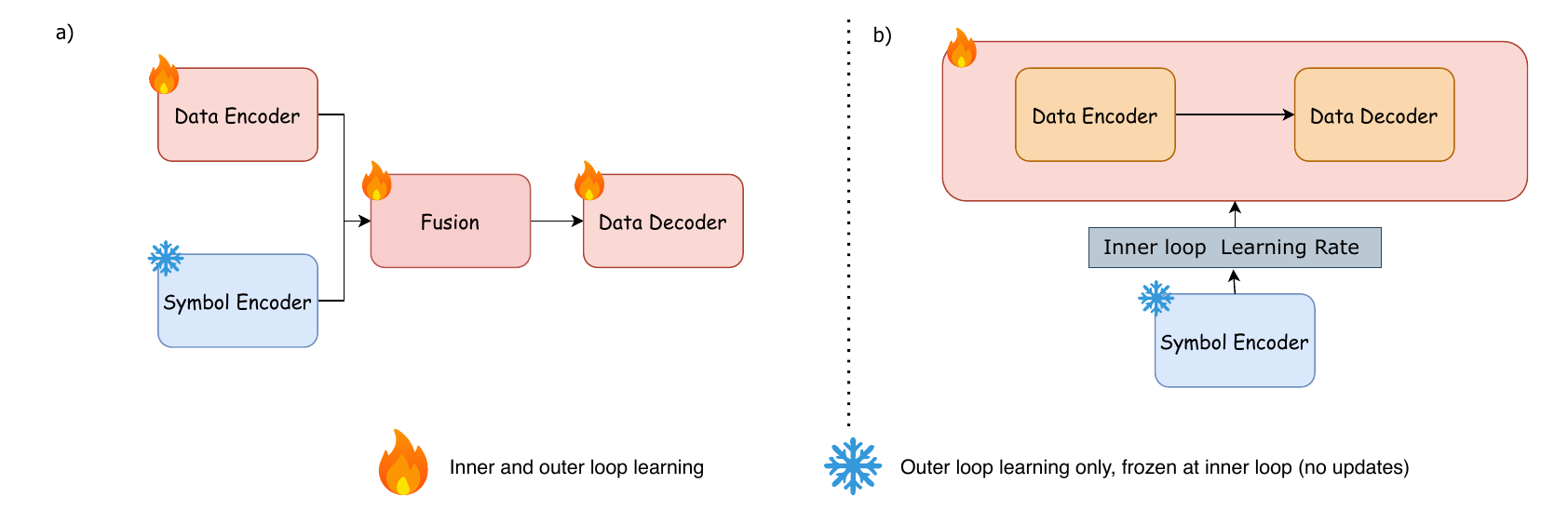}}
			\caption{\textbf{Comparison of LeMON algorithms}. a) LeMON algorithm applied on a transformer-based model, data encoding and symbol encoding are fused together by fusion layers. The symbol encoder is frozen during the inner loop updates. b) LeMON algorithm applied on a transformer-based model with learnable inner loop adaptation rate by symbolic encoder (LeMONS).  The base model here for inner loop updates is modular and can be replaced; in this work, we tested DeepONet and FNO as alternatives.}
			\label{fig:comparison_prose_lemon}
		\end{center}
		\vskip -0.2in
	\end{figure}

	The fusion layers (Figure~\ref{fig:comparison_prose_lemon}a) allow the high-efficiency information exchanges between encoded data and encoded symbolic feature, this leads the research of PROSE \cite{liu2024prose, sun2025towards}.
	However, the fusion of data and operator encoding might not be suitable for non-transformer-based models such as the popular DeepONet and FNO. 
	However, we can still incorporate symbolic encoding in our LeMON training process. As illustrated in Figure~\ref{fig:comparison_prose_lemon}b and detailed in Algorithm~\ref{algo:LeMON2}, we extend LeMON to LeMONS (LeMON with a Symbolic-Aware Adaptation Rate) by introducing a symbolic encoder $ \mathcal{M}_{\text{sym}} $ that generates operator-specific adaptation rates $ \eta_i $, conditioned on the symbolic representation $ S_i $ of $ \mathcal{G}_i $. This allows LeMONS to tailor adaptation dynamics to individual operators while jointly optimizing $ \eta_i $-generation alongside base model parameters during meta-training.
	
	For PDE $ \mathcal{G}_i $ with symbolic encoding $ S_i $, the adaptation rate becomes:
	\begin{equation}
	\eta_i = \mathcal{M}_{\text{sym}}(S_i; \theta_{\text{sym}})
	\end{equation}
	where $ \mathcal{M}_{\text{sym}} $ is a symbolic encoder network trained jointly with base model parameters $ \theta $. This enables operator-specific adaptation policies while maintaining architecture agnosticism.
	
	The symbolic module parameter $\theta_{sym}$ is updated in the outer loop (Line 15 of Algorithm \ref{algo:LeMON2}).
	Specifically, through the chain rule, similar to Eq.~\ref{chain_1}, with the first order approximation,
	we can calculate the gradients as follows: \begin{align}
	\nabla_{\theta_{sym}} \mathcal{L}_{\mathcal{G}_i}^\mathcal{M}(\mathcal{D}_i';\theta_{i,p};\theta,\theta_{sym}) &=    \nabla_{\theta_{i,p}} \mathcal{L}_{\mathcal{G}_i}^\mathcal{M}(\mathcal{D}_i';\theta_{i,p};\theta,\theta_{sym})\nabla_{\eta_i}\theta_{i,p}\nabla_{\theta_{sym}}\eta_i\\
	\nabla_{\eta_i}\theta_{i,p}&=\nabla_{\eta_i}\theta_{i,p-1}-\Delta_{p} -\eta_i \nabla_{\eta_i}\Delta_j \label{fixedeta}\\
	&\approx\nabla_{\eta_i}\theta_{i,p-1}-\Delta_{p} \\
	&\approx\cdots\approx-\sum_{j=1}^p \Delta_j\\
	\nabla_{\theta_{sym}}\eta_i&=  \nabla_{\theta_{sym}}\mathcal{M}_{sym}(S_i;\theta_{sym})
	\end{align}
	
	\begin{algorithm}[H]
		\caption{LeMON with Symbolic Adaptation Rates (LeMONS)}\label{algo:LeMON2}
		\textbf{Input}: Operator batch size $N$, inner loop step $p$, a base model $\mathcal{M}$\\
		\textbf{Parameters}: Randomly initialized network parameters $\theta$, symbolic module parameters $\theta_{sym}$ and meta update learning rate $\eta_m$\\
		\textbf{Output}: Updated network parameters $\theta$
		
		\begin{algorithmic}[1]
			\FOR{$k = 1$ {\bfseries to} End}
			\STATE Sample a batch of data associated to $N$ PDEs $\{\mathcal{G}_i(u; q_i)\}_{i=1}^N$
			\FOR{$i = 1,2,\cdots, N$}
			\STATE Obtain the symbolic information $S_i$ for the PDE  $\mathcal{G}_i(u; q_i)$
			\STATE Encode the adaptation rate $\eta_i$ by symbolic encoding module. $\eta_i = \mathcal{M}_{sym}(S_i;\theta_{sym})$
			\STATE Obtain support dataset $D_i$ by Algorithm~\ref{algo:datasampling} within  $\mathcal{G}_i(u; q_i)$ for local update
			
			\STATE Set $\theta_{i,0} = \theta$
			\FOR{$j = 1,2,\cdots, p$}
			\STATE $\Delta_j = \nabla_{\theta_{i,j-1}}\mathcal{L}_{\mathcal{G}_i}^{\mathcal{M}}(D_i; \theta_{i,j-1};\theta,\theta_{sym})$ using $D_i$ 
			\STATE$\theta_{i,j} = \theta_{i,j-1} - \eta_i \Delta_j$
			\ENDFOR
			\STATE Obtain query dataset $D_i'$ by Algorithm~\ref{algo:datasampling} with  $\mathcal{G}_i(u; q_i)$ for meta update
			\ENDFOR
			\STATE $\theta \leftarrow \theta - \eta_m \sum_{i} \nabla_{\theta} \mathcal{L}_{\mathcal{G}_i}^\mathcal{M}(\mathcal{D}_i';\theta_{i,p};\theta,\theta_{sym})$
			\STATE $\theta_{sym} \leftarrow \theta_{sym} - \eta_m \sum_{i} \nabla_{\theta_{sym}} \mathcal{L}_{\mathcal{G}_i}^\mathcal{M}(\mathcal{D}_i';\theta_{i,p};\theta,\theta_{sym})$
			\ENDFOR
		\end{algorithmic}
	\end{algorithm}

	\subsection{Implementation Details}\label{subsec:implementation}
	The base architecture $ \mathcal{M} $ can instantiate a wide variety of DNO, including DeepONet, FNO, or transformer-based methods. The symbolic encoding $ S_i $ usea equation tokenization from~\cite{liu2024prose}. Training alternates between operator batches (Algorithm~\ref{algo:LeMON1}) and symbolic rate adaptation (Algorithm~\ref{algo:LeMON2}), with experimental configurations detailed in Section~\ref{sec_exp}.

	\section{Numerical Experiments}\label{sec_exp}
	
	We study various examples of learning-to-learn PDE solution operators, i.e. solving PDE $\mathcal{G}(u; q)$. 
	More specifically, we investigate the operator $G$ mapping initial samples $u(t_0)$ to the solutions at future time-steps $u(\tau_0), ..., u(\tau_{N_{\tau}})$.
	We consider 19 families of PDEs, detailed in Appendix Table \ref{table:Notations}. Please refer to Table \ref{tab:model_hyper} for model details, and Table \ref{tab:optim_hyper} for training details.
	Unless otherwise specified, the error is reported as the relative $L^2$ error:
	$$\frac{1}{N_{\text{test}}}\sum_{i=1}^{N_{\text{test}}}\frac{\|u_{\text{pred}}^i - u_{\text{target}}^i\|_2}{\|u_{\text{target}}^i\|_2+\varepsilon},$$ 
	where the summation indexed by $i$ is taken over a total of $N_{\text{test} } =3.2 \text{K}\times N_{\text{family}}$ testing samples, and $N_{\text{family}}$ is the number of PDE-family.
	For reproducing, all codes and numerical results are available on GitHub.

	\subsection{Symbolic Module helps with Limited PDE Task-Based Data}
	We will demonstrate the significance of data diversity in producing an effective pre-trained model for downstream tasks. To this end, we consider four representative models: two SOL models—DeepONet and FNO—a transformer-based model without operator encoding (which serves as the base architecture for many popular MOL models), and PROSE, a MOL model equipped with symbolic operator encoding. Each model is pre-trained on data from various PDEs and subsequently fine-tuned using a very limited amount of target data which would be insufficient to train a model from scratch. We then compare their prediction accuracies to assess the impact of data diversity in pre-training.
	
	In this example, our goal is to approximate the solution operator of the Sine-Gordon equation. We pre-train models, DeepONet, FNO, Transformer-based, and PROSE, using 81K samples corresponding to solution operators of different PDEs. After pre-training, we fine-tune the models on 100 data samples from Sine-Gordon PDE solution operators. The detailed PDE and settings are in Appendix~\ref{sec:Notation}. 
	We consider the following pre-training datasets settings.
	\begin{itemize}
		\item \textbf{Klein-Gordon Equation}:  This family is structurally closest to the Sine-Gordon equation (with respect to the set of PDE considered), with the general form $u_{tt} + q\sin(u) = pu_{xx}$. When $p\approx 1$ and the initial condition $u_0$ is normalized to $[0,1]$, the primary distinction between Klein-Gordon and Sine-Gordon lies in the range of the free parameter $q$. Thus, models pre-trained on Klein-Gordon are expected to transfer well to Sine-Gordon.
		\item \textbf{Wave Equation}: The wave equation, $u_{tt} = pu_{xx}$ lacks the nonlinear term, making it a more distant family compared to the Klein-Gordon equation. Variability in the coefficient $p$ further differentiates it from the Sine-Gordon equation.
		\item \textbf{18 families}: This dataset consists of all other operator families available in the dataset listed in Appendix~\ref{sec:Notation}, with 4.5K data points per family. 
		\item \textbf{None}: As a baseline, we train models from scratch (the model is randomly initialized) using the same amount of Sine-Gordon data samples as used to the previous experiments, with no pre-training involved.
	\end{itemize}
	
	\begin{table}[H]
		\small
		\centering
		\setlength{\tabcolsep}{4pt}
		\renewcommand{\arraystretch}{1.5} 
		
		\begin{tabular}{c|c|cccc}
			\hline
			\textbf{Pre-training}& \multirow{2}{*}{\textbf{FT?}}&\multicolumn{4}{c}{\textbf{Model}}\\\cline{3-6}
			\textbf{ Operator Family(s)} && \textbf{DeepONet} & \textbf{FNO} & \textbf{Transformer} & \textbf{PROSE} \\ \hline\multirow{2}{*}{Klein-Gordon} &\xmark&19.6\%&19.5\%&19.4\%&20.1\%\\
			&\cmark& 18.5\%&2.43\%&2.98\%&3.07\%\\\hline
			\multirow{2}{*}{Wave} &\xmark&65.9\%&65.9\%&66.2\%&65.6\%\\
			&\cmark&30.3\%&32.2\%&10.0\%&5.92\%\\\hline
			\multirow{2}{*}{18 Families}&\xmark&50.0\%&65.3\%&38.1\%&59.4\%\\
			&\cmark&19.0\%&*&5.70\%&2.75\%\\\hline
			None&\cmark&37.5\%&36.1\%&38.0\%&25.2\%\\
			
		\end{tabular}
		\caption{\textbf{Fine-tune with Sine-Gordon}: Rel-$L^2$ error with Sine-Gordon operators. FT: Fine-tune by 100 
			data within Sine-Gordon family with 100 epochs. MOL benefits from multiple data sources as seen by the 18 families row. *Failed to converge.}
		\label{table:Scaling}
	\end{table}
	
	We present the numerical results in Table \ref{table:Scaling}. Since each neural operator has a different number of trainable parameters, thus we focus the analysis of the each column separately.
	\begin{enumerate}
		\item First, due to the limited availability of target Sine-Gordon data, none of the models achieve satisfactory prediction accuracy when trained from random initialization (as shown in the last row of the table). This highlights the importance of addressing new PDE problems with limited data by leveraging a pre-trained model.
		\item PROSE, a MOL model with symbolic operator encoding capable of handling data from multiple operators, is shown to be an effective initialization, achieving a relative error of $2.75\%$ when pre-trained on diverse families of PDE operators. This holds even if some exhibit behaviors markedly different from the target Sine-Gordon PDE. Note that fine-tuning this model yields the best prediction accuracy, outperforming even the model pre-trained on Klein-Gordon data, which is more similar to the target PDE.
		\item For the transformer-based model, which does not include operator encoding, the model pre-trained on Klein-Gordon data achieves the highest prediction accuracy, while the model pre-trained on wave equation data performs poorly. This highlights the uncertainty and limitations of relying on a single-operator dataset for pre-training, as it is often unclear which operator data most closely resembles the target, and such data may not be available in practice.
		In contrast, the model pre-trained on diverse operator data achieves a satisfactory error. Taken together with the results from the PROSE model, these findings demonstrates the importance of data diversity in obtaining a robust pre-trained model for downstream adaptation in data-scarce scenarios.
		\item While PROSE and transformer-based MOL models (without symbolic operator encoding) achieve comparable results when pre-trained solely on Klein-Gordon data, PROSE consistently outperforms in all other scenarios. This demonstrates that symbolic operator encoding enhances the generalization capability of MOL. Additional numerical results highlight how the symbolic module can be used to generate learnable learning rates and further improve accuracy (see Section~\ref{num_Lemon}).
	\end{enumerate}

	\subsection{Testing unseen PDE and Comparisons to Standard Fine-Tuning}\label{num_Lemon}

	In the previous section, we demonstrated that data diversity plays a critical role in generating effective initializations for downstream tasks. However, the choice of pre-training and subsequent fine-tuning strategies is equally important for achieving both strong initialization and efficient adaptation.
	In this section, we validate the effectiveness of our proposed learning-to-learn algorithm (Algorithm \ref{algo:LeMON1}) by comparing its performance with standard pre-training and fine-tuning methods. Furthermore, we evaluate Algorithm~\ref{algo:LeMON2}, which leverages symbolic operator encoding to learn the inner-loop learning rate (Line 15 of Algorithm~\ref{algo:LeMON2}) for adaptation, by applying the algorithms to the three popular frameworks. The results confirm that symbolic encoding significantly enhances the performance of MOL models. 
	Notably, the standard SOL frameworks evaluated in our study are inherently designed to approximate one operator at a time, which conflicts with the principle of data diversity. 
	In contrast, our proposed algorithm enables SOL architectures to be trained and fine-tuned using data from multiple operators, thereby transforming them into effective MOL frameworks.
	
\begin{figure}[htbp]
		\scriptsize
		\centering
		\begin{subfigure}{0.45\textwidth}
			\centering
			\includegraphics[width=\linewidth]{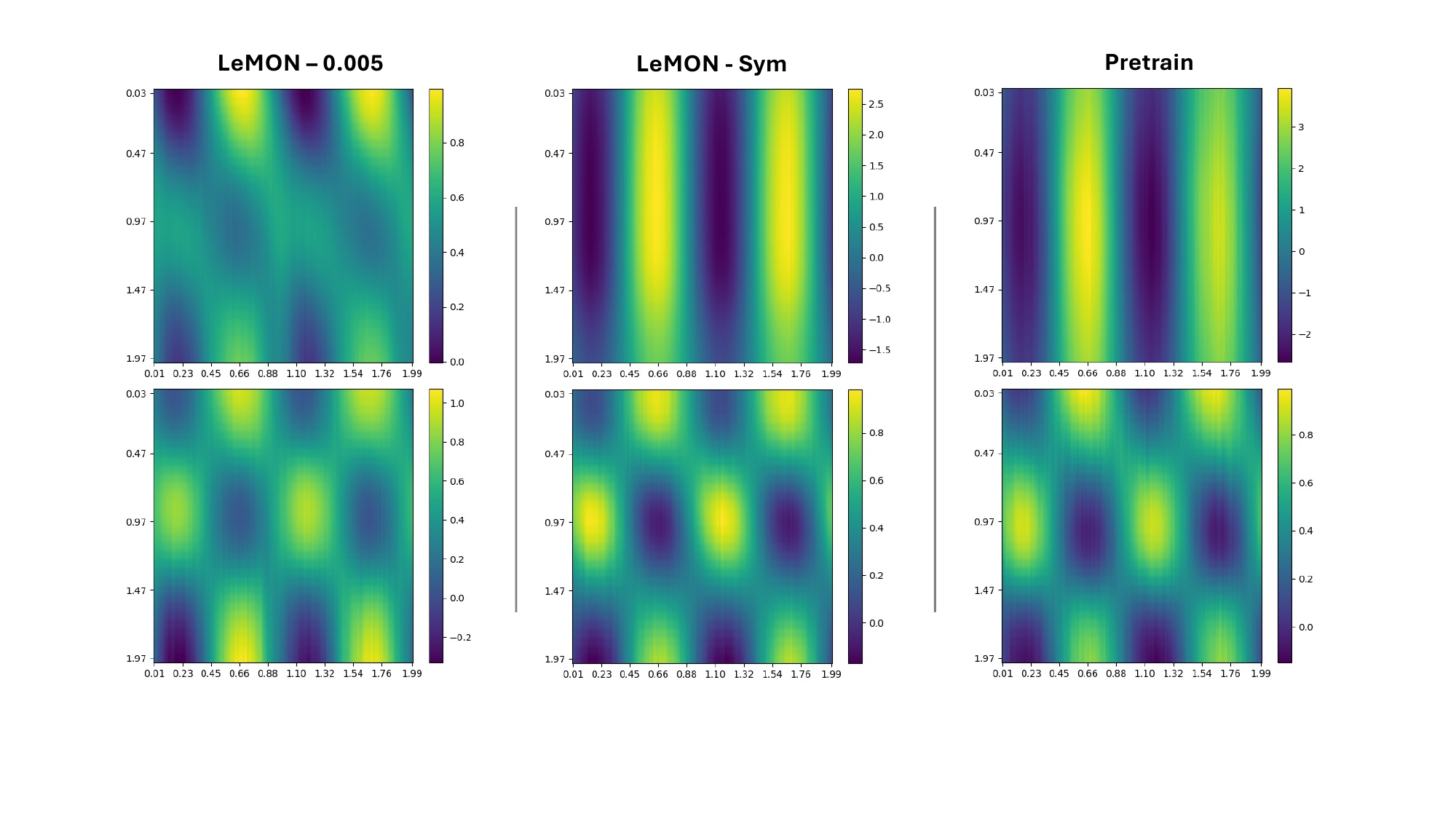}
			\vspace{-1cm}
			\caption{{\tiny\textbf{DeepONet}}:{\tiny$\geq100\%\to 8.75\%|\geq100\% \to 8.64\%|39.5\% \to 20.4\%$} }
			\label{fig:a}
		\end{subfigure}
		\begin{subfigure}{0.45\textwidth}
			\centering
			\includegraphics[width=\linewidth]{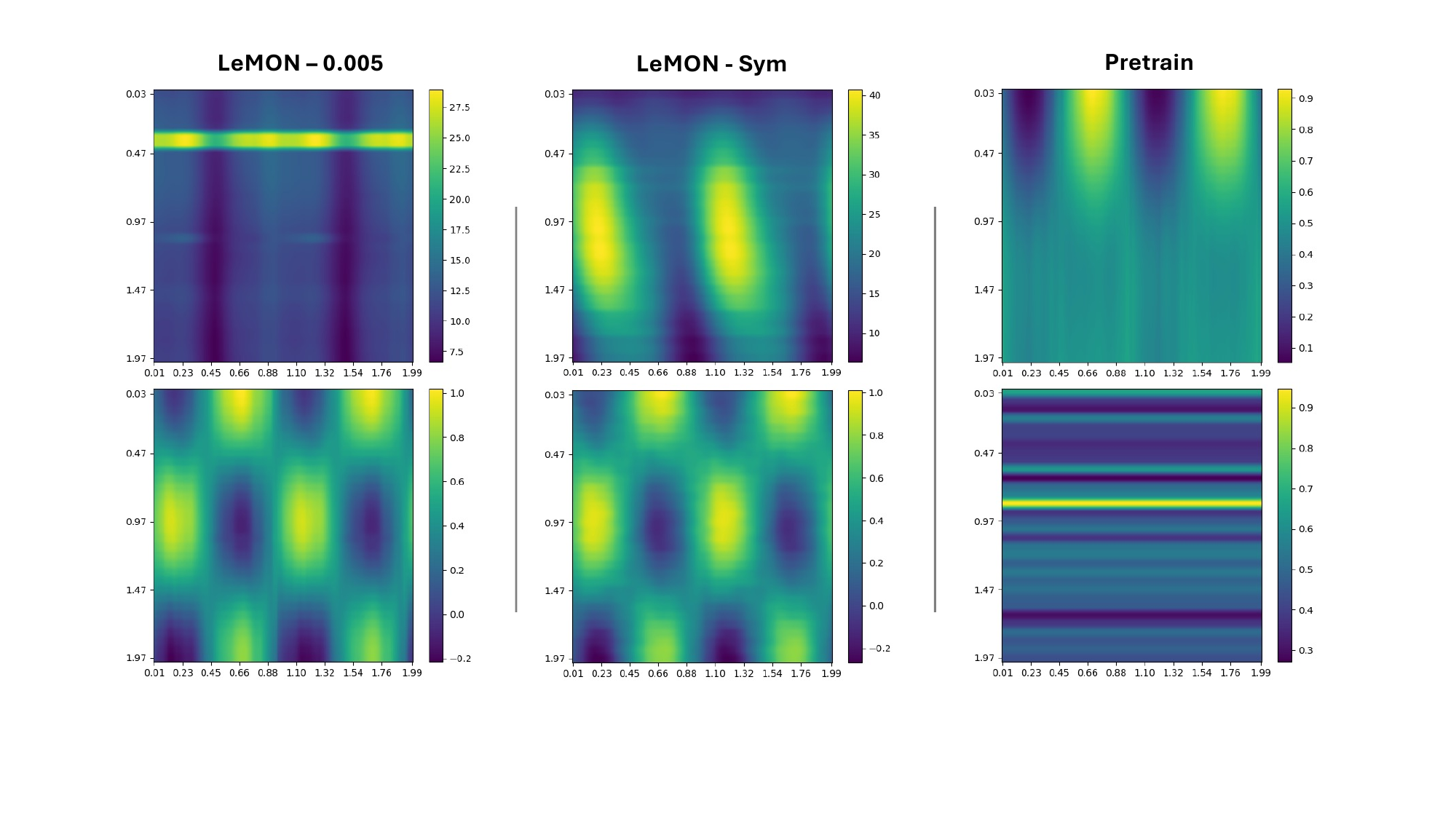}
			\vspace{-1cm}
			\caption{{\tiny\textbf{FNO}}:{\tiny$\geq100\%\to 4.72\%|\geq100\% \to 5.90\%|54.8\% \to * $} }
			\label{fig:b}
		\end{subfigure}
		
		\begin{subfigure}{0.45\textwidth}
			\centering
			\includegraphics[width=\linewidth]{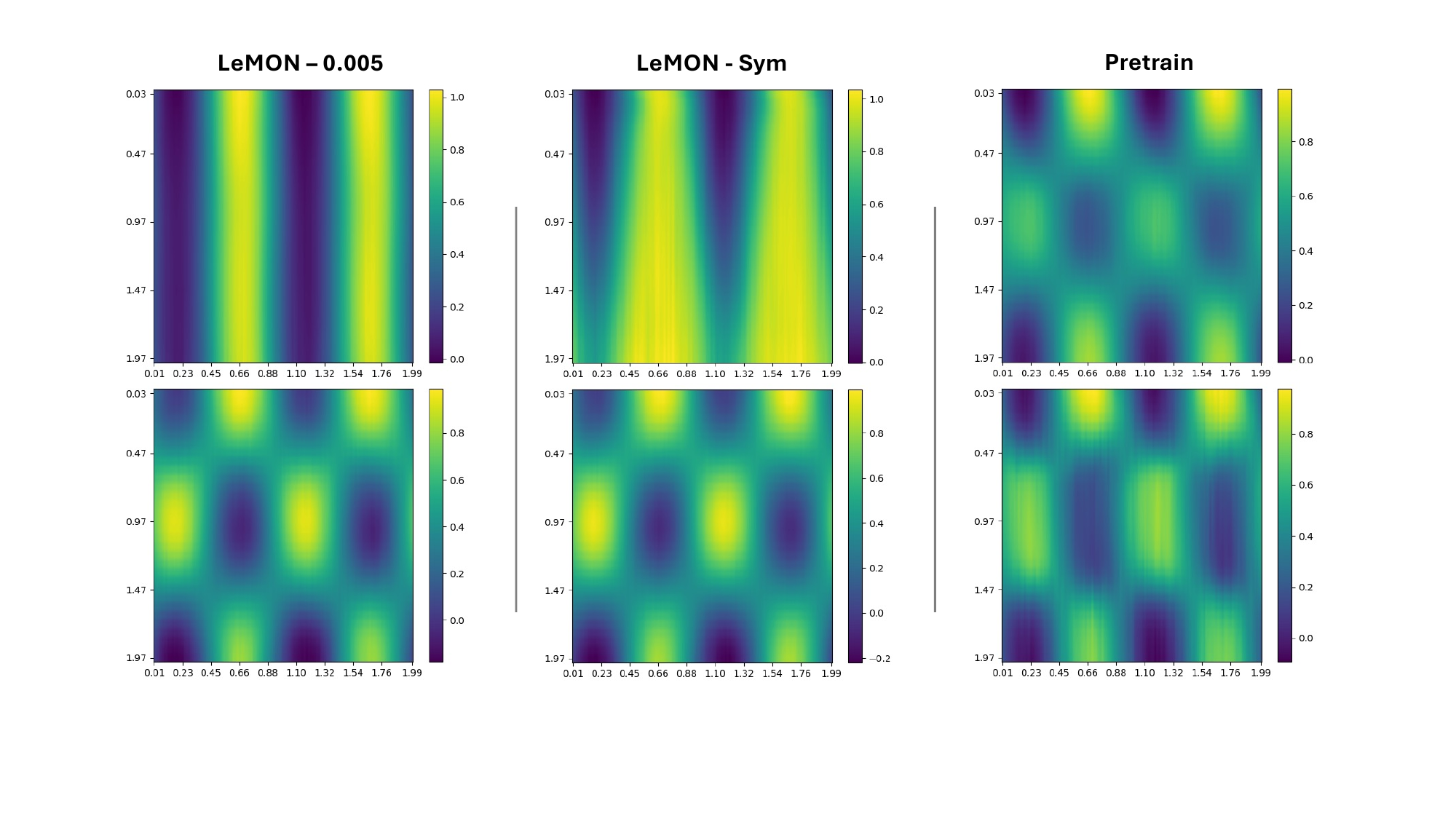}
			\vspace{-1cm}
			\caption{{\tiny\textbf{Transformer}}:{\tiny$84.4\%\to 2.15\%|95.7\% \to 1.76\%|26.9\% \to 13.5\%$} }
			
			\label{fig:c}
		\end{subfigure}
		\begin{subfigure}{0.45\textwidth}
			\centering
			\includegraphics[width=\linewidth]{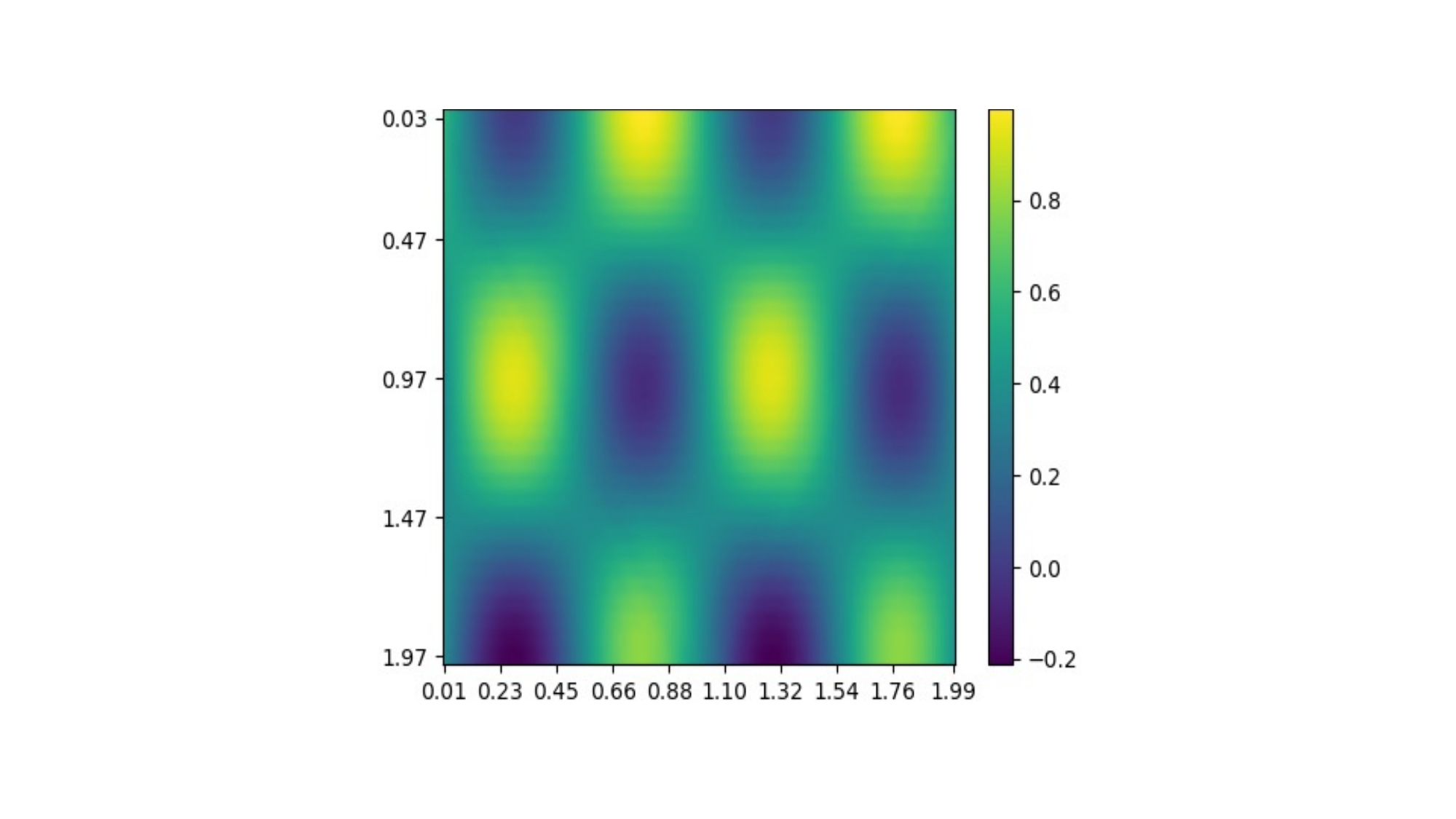}
			\vspace{-1cm}
			\caption{\textbf{Target}}
			\label{fig:d}
		\end{subfigure}
		\caption{\textbf{Sample outputs from three models}: In each of the first three figures, the first column shows results from LeMON training with a fixed inner-loop adaptation rate of 0.005 (LeMON-0.005), the second column presents results using a learnable inner-loop via the symbolic module (LeMONS), and the third column corresponds to standard pre-training (pre-train) with an inner-loop adaptation rate of 0.005 during evaluation for comparison.  The first row displays results before inner-loop adaptation, while the second row shows results after adaptation. The caption of these three figures indicates the method and the relative $L^2$ error for this specific instance before and after inner-loop adaptations under three different training methods: LeMON-0.005, LeMONS and pre-train, respectively. The final figure is the target for reference. }
		
		\label{fig:samples}
	\end{figure}

	Specifically, we consider four models in our study: DeepONet, FNO, a Transformer-based model without operator encoding, and (2-to-1) PROSE. The PROSE model is a transformer-based and integrates symbolic encoding and data encoding through fusion layers, as illustrated in Figure~\ref{fig:comparison_prose_lemon}. A representative example of this architecture is the recently proposed PROSE framework~\cite{liu2024prose, sun2025towards}. We present the results in Table \ref{table:maml_base} and Figure \ref{fig:samples}.
	
	Three SOL models are meta-trained using Algorithm~\ref{algo:LeMON1} and \ref{algo:LeMON2} on 15 operator families, each comprising 90 operators with 50 samples per operator. During training, inner-loop adaptation uses 5 gradient steps. For evaluation, models adapt to unseen operators via 20 gradient steps on 10 task-specific samples, with performance measured on 40 held-out samples. For more data detailed setup, we refer to Appendix~\ref{sec:Notation}. We test fixed adaptation rates (0.1 to 0.005; Algorithm~\ref{algo:LeMON1}) and symbolically conditioned rates (Algorithm~\ref{algo:LeMON2}). To ensure fairness, symbolic encodings in PROSE remain frozen during inner-loop updates, preserving their task-specific representation.

	\begin{table}[H]
		\centering
		\small
		\setlength{\tabcolsep}{4pt}
		\renewcommand{\arraystretch}{1.5} 
		\begin{tabular}{c|c|c|c|c|c|c}
			\hline
			\multirow{2}{*}{\textbf{Train Methods}} &\multirow{2}{*}{\textbf{Adaptation rate}}& \multirow{2}{*}{\textbf{FT?}}&\multicolumn{4}{c}{\textbf{Operator Learning Framework}}\\\cline{4-7}
			&&&\textbf{DeepONet}&\textbf{FNO} &\textbf{Transformer} & \textbf{PROSE}\\ \midrule
			\multirow{8}{*}{LeMON}&\multirow{2}{*}{0.1}&\xmark&-&-&$\geq 100$\%&90.6\%\\
			&&\cmark&-&-&11.4\%&6.18\%\\\cline{2-7}
			&\multirow{2}{*}{0.05}&\xmark&-&$\geq 100$\%&40.0\%&27.0\%\\
			&&\cmark&-&20.0\%&4.71\%&\textbf{2.61\%}\\\cline{2-7}
			&\multirow{2}{*}{0.005}&\xmark&$\geq 100$\%&$\geq 100$\%&41.2\%&7.53\%\\
			&&\cmark&10.1\%&14.2\%&\textbf{4.00\%}&2.32\%\\\cline{2-7}
			&\multirow{2}{*}{0.0005}&\xmark&$\geq 100$\%&48.5\%&35.2\%&2.78\%\\
			&&\cmark&11.3\%&17.1\%&12.2\%&2.77\%\\\hline
			\multirow{2}{*}{LeMONS}&\multirow{2}{*}{Learn}&\xmark&$\geq 100$\%&45.5\%&52.8\%&-\\
			&&\cmark&\textbf{9.43\%}&\textbf{12.1\%}&\textbf{4.00\%}&-\\
			\hline
			\multirow{2}{*}{pre-train}&\multirow{2}{*}{N/A (0.005 for eval)}&\xmark&28.3\%&32.8\%&\multicolumn{2}{c}{24.9\%}\\
			&&\cmark&20.2\%&*&\multicolumn{2}{c}{24.5\%}\\\hline
		\end{tabular}
		\caption{\textbf{Comparison between LeMON with different inner-loop adaptation rate and standard pre-training method.} Five-step gradient descent is used during LeMON training, and twenty-step gradient descent during all evaluation. For PROSE model, the symbolic encoding module is not updating during LeMON training. FT (Fine-tuning)?: \xmark/ \cmark performance before/after inner-loop updates. Results are $L^2$ errors.  DeepONet, FNO, Transformer: SOL architectures. PROSE: Transformer-based model, symbol encoding and data encoding are fused together by fusion layers as illustrated in Figure~\ref{fig:comparison_prose_lemon}. *Fail to converge. }
		\label{table:maml_base}
	\end{table}

	Since the models in this comparison have different numbers of trainable parameters, we analyze the results column-wise in the table. That is, we do not compare between the difference architectures, but rather, compare the advantages of including LeMON and LeMONS to the different neural operators. To ensure a fair comparison, we have optimized the number of trainable parameters DeepONet and FNO models up to the amount where further increases in then number of trainable parameters no longer yield reductions in prediction error.
	We have the following observations from the Table \ref{table:maml_base}.
	\begin{enumerate}
		\item First, the proposed learning-to-learn pre-training and fine-tuning strategies outperform standard training and fine-tuning methods (i.e., AdamW for both stages) across all models, as shown in the results presented in the last row of the table.
		\item 
		The performance of the method is sensitive to the choice of the inner-loop learning rate (referred to as the adaptation rate in the table). For instance, in the case of DeepONet, the post-adaptation error, i.e., the prediction error after fine-tuning on operator-specific data, significantly deteriorates when the learning rate is relatively large ($\geq 0.05$), leading to instability.
		Our proposed strategy for learning the inner-loop learning rate via symbolic encoding yields consistently superior results across all models. In particular, within each model column, the learnable adaptation rate obtained through symbolic encoding achieves the lowest post-adaptation errors.    
	\end{enumerate}
	
\begin{figure}[H]
		\scriptsize
		\centering
		\begin{subfigure}{0.45\textwidth}
			\centering
			\includegraphics[width=\linewidth]{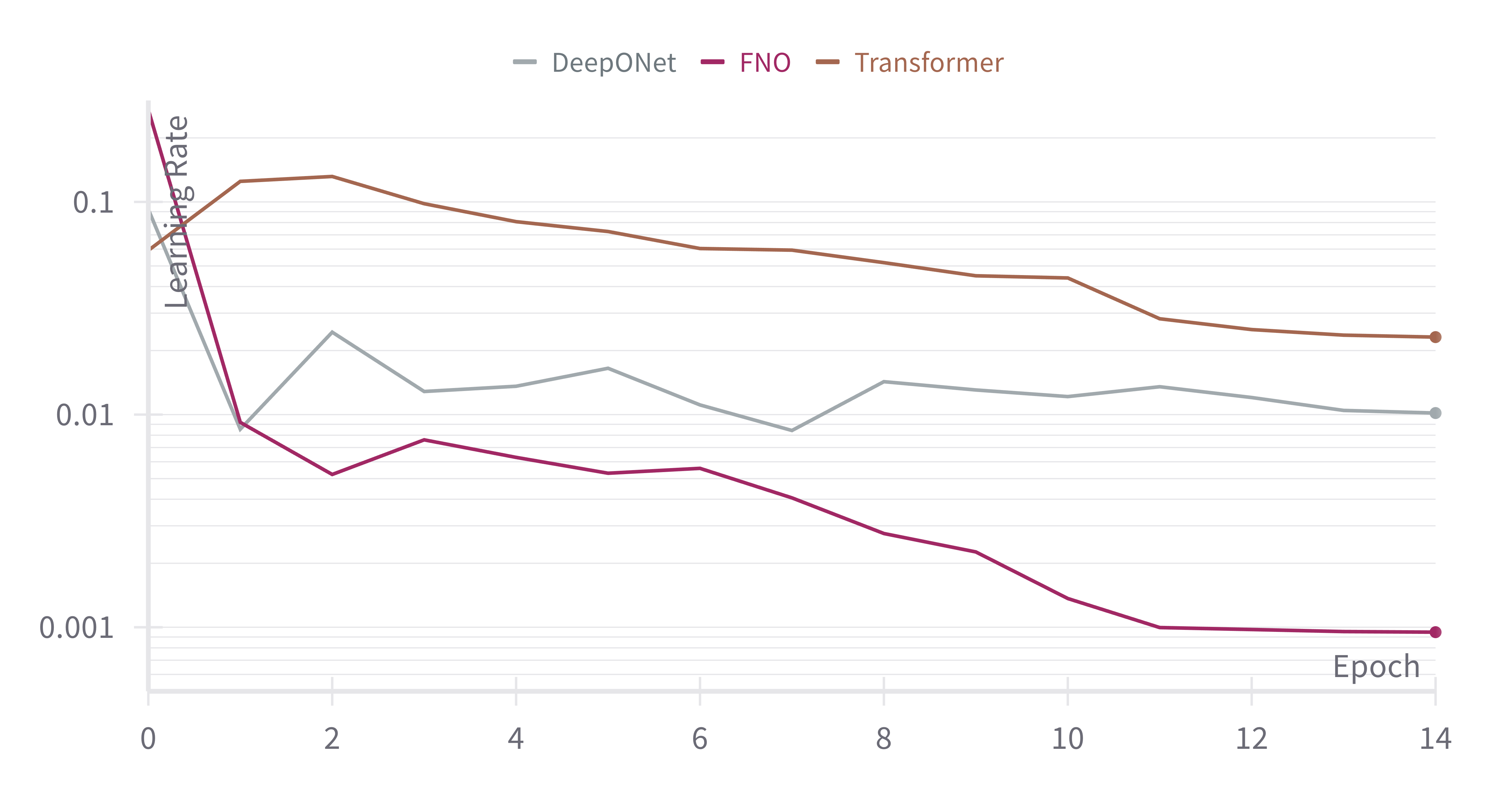}
			\vspace{-.5cm}
			\caption{Klein-Gordon}
			\label{fig:a1}
		\end{subfigure}
		\begin{subfigure}{0.45\textwidth}
			\centering
			\includegraphics[width=\linewidth]{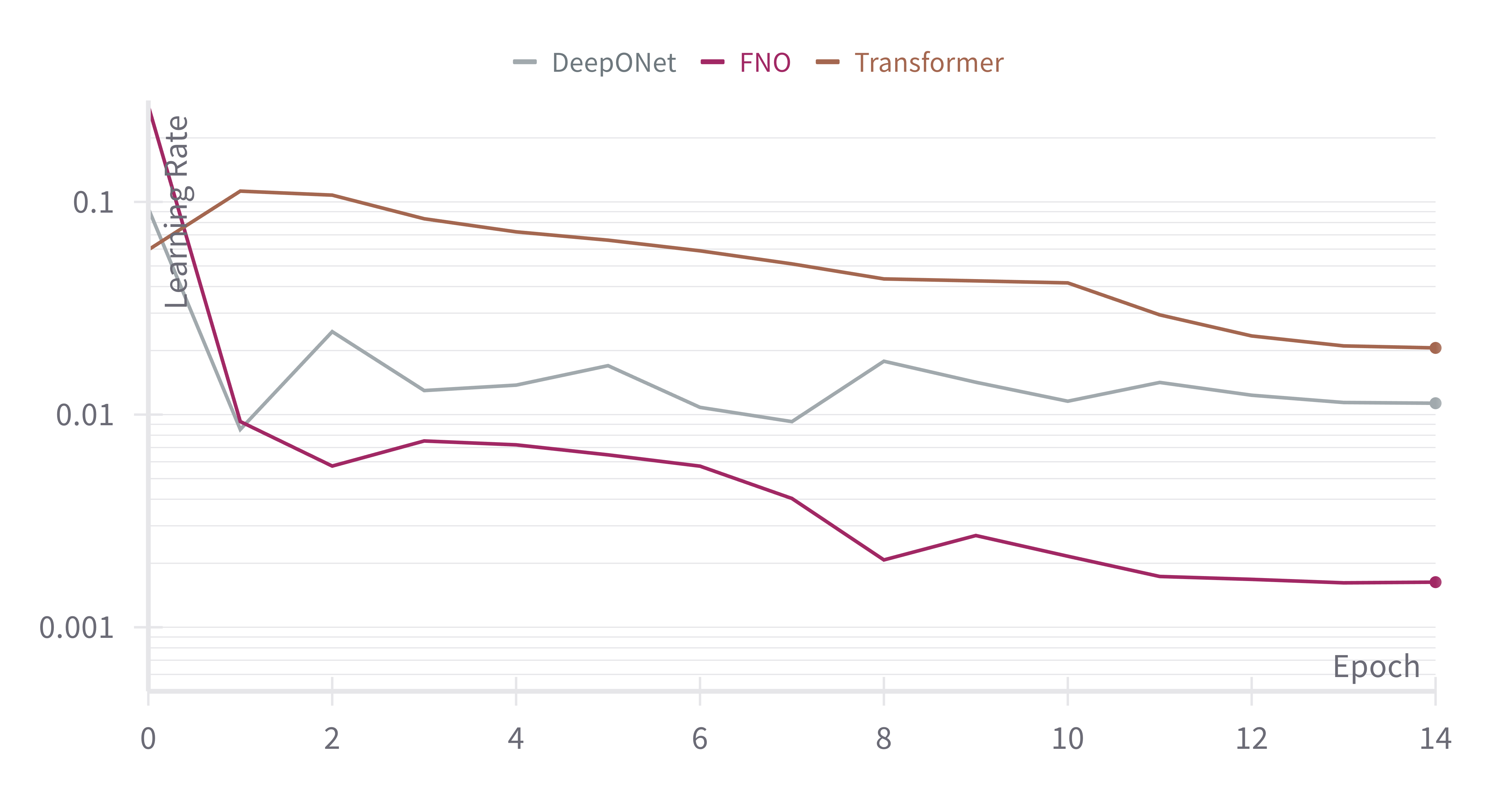}
			\vspace{-.5cm}
			\caption{Porous Medium}
			\label{fig:b1}
		\end{subfigure}
		
		\begin{subfigure}{0.45\textwidth}
			\centering
			\includegraphics[width=\linewidth]{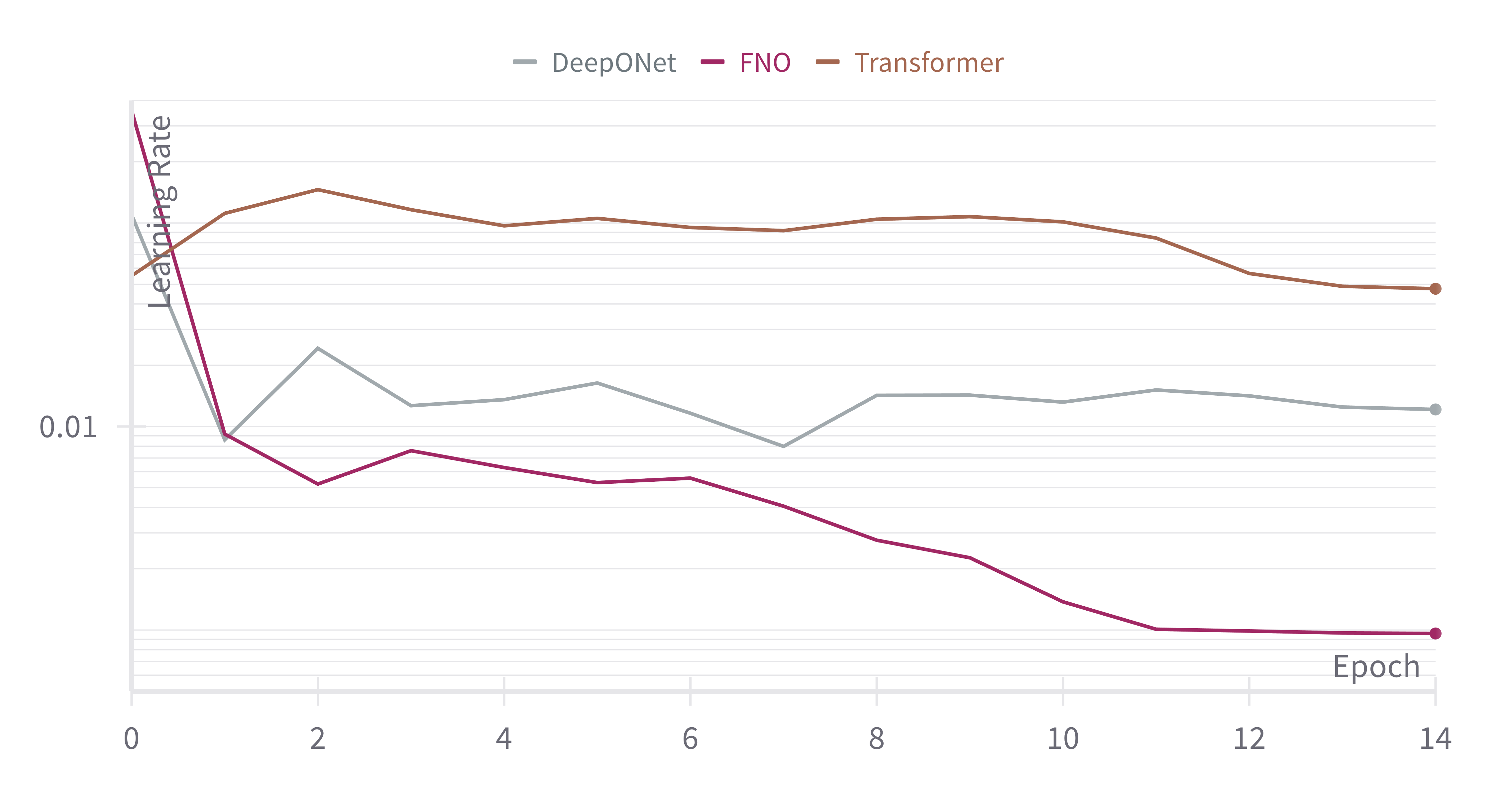}
			\vspace{-.5cm}
			\caption{Wave}
			
			\label{fig:c1}
		\end{subfigure}
		\begin{subfigure}{0.45\textwidth}
			\centering
			\includegraphics[width=\linewidth]{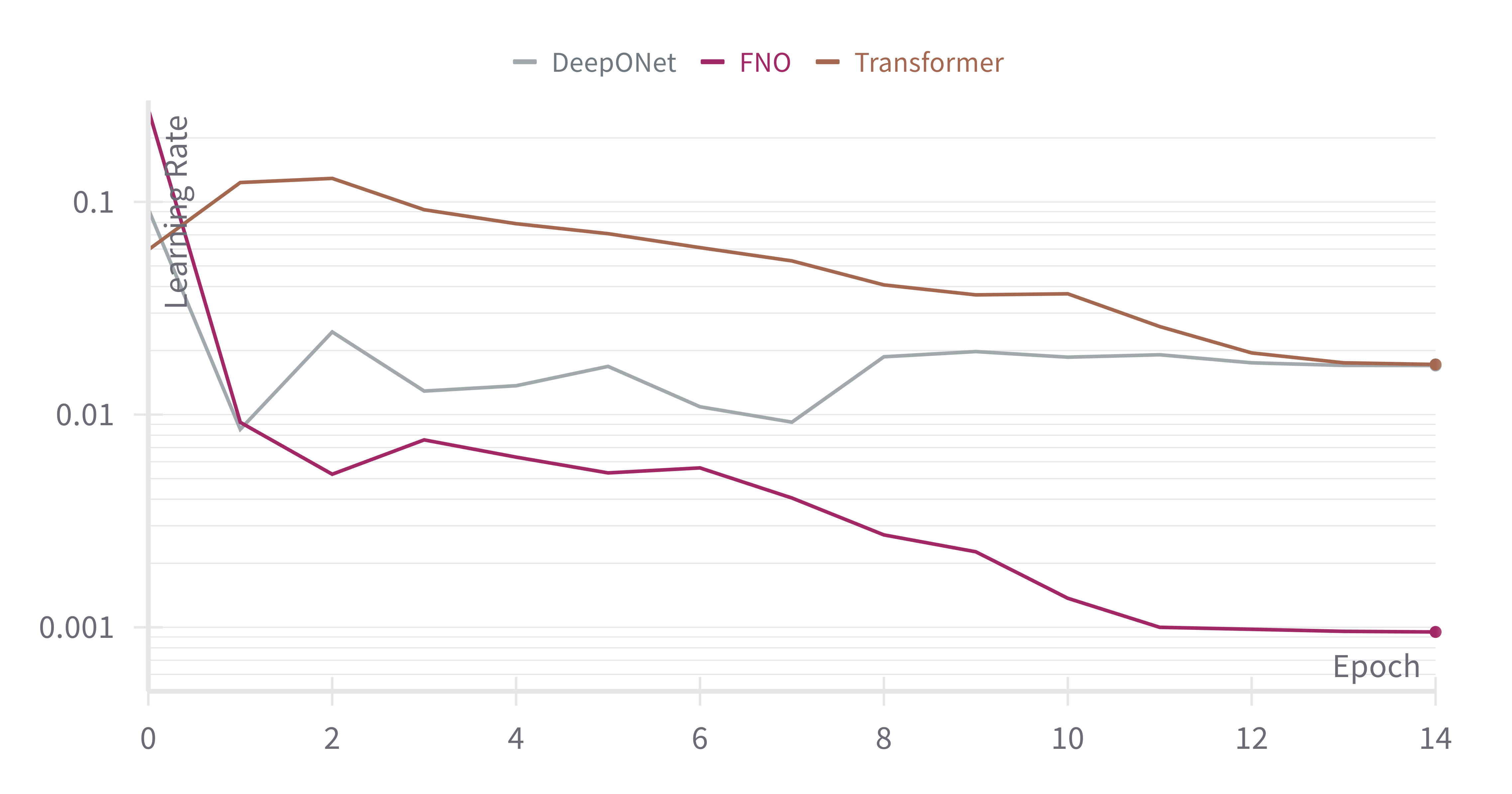}
			\vspace{-.5cm}
			\caption{Burgers'}
			\label{fig:d1}
		\end{subfigure}
		\caption{\textbf{Trend of Learning rate in LeMONS}: We present one instance from each of the four operator families: Klein-Gordon, Porous Medium, Wave, and Burgers', and show the trend of learned learning rate during the training process of LeMONS. }
		
		\label{fig:lrs}
	\end{figure}

	In addition to Table~\ref{table:maml_base}, we present the learned learning rates to illustrate the Algorithm \ref{algo:LeMON2}. See Figure~\ref{fig:lrs} for a visual demonstration.
	
	\subsection{Testing with the unseen PDE families}
	
	In this section, we evaluate the extrapolation capabilities of LeMON-trained models when applied to entirely unseen operator families. 
	Table~\ref{table:extrapolation} compares the $L^2$ errors of DeepONet, FNO, and Transformer architectures under three training paradigms: LeMON (fixed adaptation rate), LeMONS (symbolic-conditioned adaptation rate), and traditional pre-training (no adaptation during training). 
	Models are meta-trained on 13 operator families (Appendix~\ref{sec:Notation}) and tested on four held-out operator families (Burgers, Klein-Gordon, Diff-Reaction, and Diffusion).
	
	LeMONS consistently outperforms LeMON and pre-training across all models, particularly after fine-tuning, but with an increased 30\% the optimization time and 10M trainable parameters. For example, DeepONet with LeMONS achieves a 5.00\% error on the Diffusion dataset post-FT, compared to 5.26\% for LeMON and 15.8\% for pre-training. This demonstrates that dynamically learned rates, conditioned on symbolic operator features, enable more efficient task-specific adaptation than fixed rates. Pre-training, with a lack of adaptation mechanism, fails to generalize to unseen operators. In contrast, LeMON-based methods with inner-loop fine-tuning reduce the error as to low as $3.90\%$ (Klein-Gordon equations with symbolic adapted learning rate). This highlights the necessity of meta-learning for extrapolation in multi-operator settings. Transformers exhibit competitive performance under both LeMON and LeMONS, likely due to their inherent flexibility in processing sequential or structured data. However, FNO has difficulty with fixed-rate LeMON (pre-FT errors $\geq 1000\%$), suggesting spectral architectures may require special adaptation policies. The performance gap between pre-FT (\xmark) and post-FT (\cmark) results emphasizes the necessity of inner-loop adaptation. For instance, FNO with LeMONS reduces its Klein-Gordon error from 48.5\% to 3.90\% after just 20-step fine-tuning, validating the framework’s ability to rapidly specialize to novel operators.

	\begin{table}[H]
		\scriptsize
		\begin{center}
			\setlength{\tabcolsep}{4pt}
			\renewcommand{\arraystretch}{1.5} 
			\begin{tabular}{c|c|c|c|c|c|c|c}
				\toprule
				\textbf{Model} & \textbf{Training Method} & \textbf{Adaptation rate} & \textbf{FT?} & \textbf{Burgers} & \textbf{Klein-Gordon} & \textbf{Diff-Reaction} & \textbf{Diffusion} \\ 
				\midrule
				\multirow{6}{*}{\textbf{DeepONet}} 
				& \multirow{2}{*}{LeMONS} & \multirow{2}{*}{Learn} & \xmark & 111\% & 145\% & 53.0\% & 63.7\% \\ 
				& & & \cmark & \textbf{20.3\%} & \textbf{6.7\%} & \textbf{5.50\%} & \textbf{5.00\%} \\ 
				\cline{2-8}
				& \multirow{2}{*}{LeMON} & \multirow{2}{*}{0.005} & \xmark & 182\% & 202\% & 123\% & 146\% \\ 
				& & & \cmark & 21.0\% & 8.1\% & 5.85\% & 5.26\% \\ 
				\cline{2-8}
				& \multirow{2}{*}{pre-training} & \multirow{2}{*}{N/A (0.005 for eval)} & \xmark & 37.5\% & 16.7\% & 24.1\% & 21.6\% \\ 
				& & & \cmark & 27.2\% & 9.48\% & 16.5\% & 15.8\% \\ 
				\midrule
				\multirow{6}{*}{\textbf{FNO}} 
				& \multirow{2}{*}{LeMONS} & \multirow{2}{*}{Learn} & \xmark & 32.48\% & 48.5\% & 38.2\% & 37.0\% \\ 
				& & & \cmark & \textbf{15.9\%} & \textbf{3.90\%} & \textbf{8.97\%} & \textbf{8.85\%} \\ 
				\cline{2-8}
				& \multirow{2}{*}{LeMON} & \multirow{2}{*}{0.005} & \xmark & $\geq 1000$\% &$\geq 1000$\%  & $\geq 1000$\%  & $\geq 1000$\%  \\ 
				& & & \cmark & 20.6\% & 8.28\% & 10.8\% & 9.7\% \\ 
				\cline{2-8}
				& \multirow{2}{*}{pre-training} & \multirow{2}{*}{N/A (0.005 for eval)} & \xmark & 36.8\% & 45.3\% & 31.6\% & 30.9\% \\ 
				& & & \cmark & * & * & * & * \\ 
				\midrule
				\multirow{6}{*}{\textbf{Transformer}} 
				& \multirow{2}{*}{LeMONS} & \multirow{2}{*}{Learn} & \xmark & 62.7\% & 74.4\% & 24.0\% & 35.4\% \\ 
				& & & \cmark & 9.11\% & \textbf{5.85\%} & 4.5\% & 3.22\% \\ 
				\cline{2-8}
				& \multirow{2}{*}{LeMON} & \multirow{2}{*}{0.005} & \xmark & 44.7\% & 59.2\% & 27.1\% & 24.0\% \\ 
				& & & \cmark & \textbf{7.65\%} & 7.56\% & \textbf{3.57\%} & \textbf{2.87\%} \\ 
				\cline{2-8}
				& \multirow{2}{*}{pre-training} & \multirow{2}{*}{N/A (0.005 for eval)} & \xmark & 38.1\% & 22.2\% & 28.8\% & 28.7\% \\ 
				& & & \cmark & 37.0\% & 36.3\% & 26.1\% & 25.7\% \\ 
				\bottomrule
			\end{tabular}
		\end{center}
		\caption{\textbf{Extrapolation performance} across models and training methods. FT (Fine-tuning): \xmark/\cmark denote performance before/after inner-loop adaptation. LeMONS uses symbolic-guided adaptation rates; pre-training lacks adaptation, and we apply an adaptation rate of $0.005$ for adaptation. Results are $L^2$ errors. *Failed to converge.}
		\label{table:extrapolation}
	\end{table}
	
	\subsection{Ablation Study}

	\begin{table}[t]
		\small
		\vskip 0.15in
		\begin{center}
			\setlength{\tabcolsep}{4pt}
			\renewcommand{\arraystretch}{1.5} 
			
			\captionsetup{justification=centering} 
			
			\begin{subtable}{0.45\textwidth}
				\centering
				\begin{tabular}{c|c|c|c}
					\toprule
					\textbf{Batch Size} & \textbf{FT?} & \textbf{$L^2$ Error} & \textbf{Time*} \\ \midrule
					\multirow{2}{*}{1} & \xmark & 180\% & \multirow{2}{*}{23s} \\ 
					& \cmark & 10.90\% & \\ \cline{1-1}\cline{2-4}
					\multirow{2}{*}{2} & \xmark & 43.30\% & \multirow{2}{*}{46s} \\ 
					& \cmark & 5.39\% & \\ \cline{1-1}\cline{2-4}
					\multirow{2}{*}{3} & \xmark & 42.90\% & \multirow{2}{*}{78s} \\ 
					& \cmark & 4.56\% & \\ \cline{1-1}\cline{2-4}
					\multirow{2}{*}{4} & \xmark & 44.30\% & \multirow{2}{*}{102s} \\ 
					& \cmark & 4.53\% & \\ \cline{1-1}\cline{2-4}
					\multirow{2}{*}{5} & \xmark & 52.80\% & \multirow{2}{*}{129s} \\ 
					& \cmark & 4.00\% & \\ 
					\bottomrule
				\end{tabular}
				\caption{\textbf{Effect of Varying Batch Size (Steps = 5)}}
				\label{table:batch_size}
			\end{subtable}
			\hfill
			\begin{subtable}{0.45\textwidth}
				\centering
				\begin{tabular}{c|c|c|c}
					\toprule
					\textbf{Steps} & \textbf{FT?} & \textbf{$L^2$ Error} & \textbf{Time*} \\ \midrule
					\multirow{2}{*}{1} & \xmark & 37.30\%& \multirow{2}{*}{36s} \\ 
					& \cmark & 6.09\%& \\ \cline{1-1}\cline{2-4}
					\multirow{2}{*}{2} & \xmark & 50.70\%& \multirow{2}{*}{54s} \\ 
					& \cmark & 5.28\%& \\ \cline{1-1}\cline{2-4}
					\multirow{2}{*}{3} & \xmark & 46.5\%& \multirow{2}{*}{90s} \\ 
					& \cmark & 4.59\%& \\ \cline{1-1}\cline{2-4}
					\multirow{2}{*}{4} & \xmark & 49.10\%& \multirow{2}{*}{109s} \\ 
					& \cmark & 4.71\%& \\ \cline{1-1}\cline{2-4}
					\multirow{2}{*}{5} & \xmark & 52.80\%& \multirow{2}{*}{129s} \\ 
					& \cmark & 4.00\%& \\ 
					\bottomrule
				\end{tabular}
				\caption{\textbf{Effect of Varying Steps (Batch Size = 5)}}
				\label{table:steps}
			\end{subtable}
			
		\end{center}
		\caption{\textbf{Comparison of different batch sizes and inner-loop update steps (Transformer-based model with learnable adaptation rate)}. 
			Batch size: The number of tasks/operators updates (inner-loop updates) before one outer-loop update.  
			Steps: The number of updates within one inner-loop update.  
			FT (Fine-tuning)?: \xmark/\cmark indicate performance before/after inner-loop updates.  
			Results are $L^2$ errors.  
			Time: The time elapsed for every 100 outer-loop updates. 30K outer-loop updates per run.}
		\label{table:ablation}
		
		\vskip -0.1in
	\end{table}
	
	In this section, we analyze the impact of different LeMON training settings on the performance of transformer-based operator learning models. The data used in this section is the same as Section~\ref{num_Lemon}. Specifically, we investigate the effects of varying task batch size and the number of inner-loop update steps.
	\begin{itemize}
		\item 
		\textbf{Task Batch Size}: Refers to the number of operators used for inner-loop updates before performing one outer-loop update.
		\item 
		\textbf{Inner-Loop Update Steps}: Refers to the number of gradient descent steps executed within a single inner-loop update.
	\end{itemize}
	Table~\ref{table:ablation} summarizes the results of our experiments. It shows that increasing the batch size consistently improves model performance. However, the performance gain diminishes as the batch size becomes larger, indicating a descending marginal benefit. Increasing the number of gradient descent steps within the inner loop shows non-monotonic performance improvement. While gains are observed with additional steps, the magnitude of these gains is relatively small compared to the impact of varying task batch size. The time complexity of varying task batch size scales almost linearly, making it a predictable factor in resource allocation. Varying the number of inner-loop update steps has a sub-linear effect on computational cost, suggesting that increasing inner-loop steps has a smaller impact on runtime compared to batch size adjustments.
	
	\section{Conclusion}\label{sec_conclusion}
	
	In this work, we proposed and developed a meta-learning approach for operator learning tasks, introducing the LeMON algorithm to address the challenges of generalizing across diverse operators. By leveraging symbolic encoding, we designed a symbol-induced module that enables task-specific adaptations through learnable inner-loop adaptation rates, providing both flexibility and robustness in handling heterogeneous data. Through comprehensive testings, we demonstrated that LeMON consistently outperforms traditional pre-training-fine-tuning methods in single-operator learning tasks and exhibits competitive performance in multi-operator learning scenarios. The ablation studies further highlighted the importance of inner-loop settings, particularly task batch size and update steps, in achieving optimal performance. Moreover, our exploration of extrapolation capabilities of neural operators showed that LeMON can allow models to effectively handle unseen operators under appropriate settings.
	One observation is that the among some transformer-based operators, PROSE model showed accurate performance in zero-shot learning due to its fused symbolic and data modules. This is leveraged in LeMON, by offering adaptability for task-specific generalization with the symbolic module.

	
	

	\section*{Acknowledgement}
	H. Schaeffer was supported in part by AFOSR MURI FA9550-21-1-0084. Z. Zhang was supported by the Department of Energy DE-SC0025440.
	
	\bibliographystyle{unsrt}
	\bibliography{references}

	\newpage
	\appendix
	\section{Data}\label{sec:Notation}
	
	\subsection{Overview}
	
	We present the abbreviation and PDE details for each operators used in the paper in Table~\ref{table:Notations}.  For results in Table~\ref{table:maml_base} and Table~\ref{table:ablation}, we train models using 15 operator families (AD, PM, KdV,  DF, WV, Sine-Gordon, Klein-Gordon, CH, DL, DLo, DS, DB, B, IB, FP). For results in Table~\ref{table:extrapolation}, we train models using 13 operator families (AD, PM, KdV, WV, Sine-Gordon, CH, DLo, DS, DB, CC, InCub, IB, FP), and test models on 4 operator families (B, Klein-Gordon, DL, DF). 
	
	During each training, we select 90 operators from each family, with 50 data in each operator. For a LeMON inner-loop, we randomly sample 10 data as support set and 20 data as query set within an operator.
	
	During each evaluation (testing), we select 80 operators distinct from those in training from each family, and with 50 data in each operator. We also randomly sample 10 data as support set, and evaluate on the rest 40 data.
	
	\begin{table}[t]
		
		\caption{Notation and Families of PDEs}
		\vskip 0.15in
		\label{table:Notations}
		\centering
		\small
		\setlength{\tabcolsep}{6pt} 
		\renewcommand{\arraystretch}{1.8} 
		\begin{tabular}{c|cc}
			\hline
			\textbf{Abbreviation} & \textbf{Type}& \textbf{Equation}\\\hline
			\textbf{AD} & Advection& $u_t+qu_x=0$, $q_c=0.5$\\
			\textbf{PM} & Porous Medium& $u_t=(u^m)_{xx}$, $m\in (2,3,4)$\\
			\textbf{KdV} & Korteweg-De Vries&$u_t + q^2u_{xxx}+uu_x=0$, $q_c = 0.022$\\
			\textbf{DF} & Diffusion & $u_t=qu_{xx}$, $q_c = 3\times 10^{-3}$\\
			\textbf{WV} & Wave & $u_{tt}=qu_{xx}$, $q_c = 0.5$\\
			\textbf{Sine-Gordon} & Sine-Gordon & $u_{tt} + q\sin(u)=u_{xx}$, $q_c = 1$\\
			\textbf{Klein-Gordon} & Klein-Gordon&$u_{tt} + p^2q^4u = q^2u_{xx}$, $q_c=1, p_c=0.1$\\
			\textbf{CH} & Cahn-Hilliard& $u_{t} + q^2u_{xxxx} + 6(uu_x)_x =0$, $q_c=0.01$\\
			\textbf{DL} & Diffusion - (Linear) Reaction& $u_t = qu_{xx}+pu$, $q_c= 3\times 10^{-3}$, $p_c = 0.1$ \\
			\textbf{DLo} & Diffusion - (Logistic) Reaction& $u_t = qu_{xx}+pu(1-u)$, $q_c= 3\times 10^{-3}$, $p_c = 1$ \\
			\textbf{DS} & Diffusion - (Square Logistic) Reaction &$u_t = qu_{xx}+pu^2(1-u)^2$, $q_c= 3\times 10^{-3}$, $p_c = 1$\\
			\textbf{DB} & Diffusion - (Bistable) Reaction &$u_t = qu_{xx}+pu^2(1-u)$, $q_c= 3\times 10^{-3}$, $p_c = 1$ \\
			\textbf{SinF}&  Conservation Law with Sine Flux& $u_t + q(\sin(u))_x = \frac{p}{\pi}u_{xx}$, $q_c=1$, $p_c=0.01$\\
			\textbf{InSin}& Inviscid Conservation Law with Sine Flux& $u_t + q(\sin(u))_x = 0$, $q_c=1$\\
			\textbf{Cos}&  Conservation Law with Cosine Flux& $u_t + q(\cos(u))_x = \frac{p}{\pi}u_{xx}$, $q_c=1$, $p_c=0.01$\\
			\textbf{InCos}& Inviscid Conservation Law with Cosine Flux& $u_t + q(\cos(u))_x = 0$, $q_c=1$\\
			\textbf{CC}& Conservation Law with Cubic Flux&$u_t + q\left(\frac{u^3}{3}\right)_x = \frac{p}{\pi}u_{xx}$, $q_c=1$, $p_c=0.01$\\
			\textbf{InCub}& Inviscid Conservation Law with Cubic Flux&$u_t + q\left(\frac{u^3}{3}\right)_x = 0$, $q_c=1$\\
			\textbf{B}& Burgers'&$u_t + q\left(\frac{u^2}{2}\right)_x = \frac{p}{\pi}u_{xx}$, $q_c=1$, $p_c=0.01$\\
			\textbf{InB}& Inviscid Burgers'&$u_t + q\left(\frac{u^2}{2}\right)_x =0$, $q_c=1$\\
			\multirow{2}{*}{\textbf{FP}} & \multirow{2}{*}{Fokker-Planck} & $u_t=Du_{xx}- \frac{(U(x)_xu)_x}{\gamma}$, $U(x) = 5\times 10^{-21}\cos(10^7x)$\\&& $D = \frac{300k_B}{\gamma}$, $\gamma=6\pi \times 10^{-7} q$, $q_c=10^{-3}$\\
		\end{tabular}
	\end{table}
	
	\subsection{Initial and boundary Conditions}
	We mainly consider periodic boundary conditions for the most of the result, and we use different types of initial conditions for different types of equations:

	\subsubsection*{Super-position of sinusoidal waves} 
	This is derived from PDEBench \cite{takamoto2022pdebench}:
	
	\begin{equation} \label{sineIC}
	u_0(x) = \sum_{k=k_1,\cdots, k_N} A_i \sin(k_ix+\phi_i)
	\end{equation}
	where $k_i = 2\pi n_i/L_x$, $n_i$ is randomly selected integers from $[1,n_{max}]$, $L_x$ is the spatial domain size. The amplitude $A_i$ is a random float uniformly chosen in $[0,1]$, and $\phi_i$ is the randomly chosen phase in $(0,2\pi)$. For all equations except advection and wave equation, after calculating \eqref{sineIC}, we enforced absolute value with random signatures and the window function with 10\% probability.
	
	\subsubsection*{Gaussian Distribution}
	\begin{equation}\label{Gaussian}
	u_0(x)= \sum_{i=1}^{N} A_i \exp \left( - \frac{|x - \mu_i|^2}{2\sigma_i^2}\right)
	\end{equation}

	\subsubsection*{Periodization and normalization}
	
	We enforced periodicity for initial condition $u_0(x)$ by removing the linear function $l(x)$ that passes through the endpoints of the domain, hence the modified initial condition is given by \begin{equation}u_0'(x) = u_0(x) - l(x).\end{equation} We also normalize the initial condition in two distinct ways:
	\begin{itemize}
		\item Adjust $u_0'(x)$ so that the range falls within $(0,u_{max})$.
		\item When the initial condition is represented by a probability distribution, we adjusted $u_0(x)$ such that the sum over all points is $1$.
	\end{itemize}
	
	\begin{table}[ht]
		\centering
		\caption{Choice of Training and Evaluation Initial Condition for different types of equations}
		\vskip 0.15in
		\label{tab:IC}
		\begin{tabular}{c|c}
			\hline
			Equation type &  Initial Condition    \\\hline
			Heat&\multirow{10}{*}{ \eqref{sineIC}: $n_{max} = 2$}\\
			Diff-React&\\
			Klein-Gordon&\\
			Sine-Gordon&\\
			Cahn-Hilliard&\\
			Viscous Conservation&\\
			Inviscid Conservation&\\ 
			Kdv&\\
			Advection&\\
			Wave&\\ \cline{2-2}
			Fokker-Planck&\multirow{2}{*}{\eqref{Gaussian}: $N=1$}\\ 
			Porous medium &\\\hline
		\end{tabular}
	\end{table}

	\subsection{Solvers}
	\label{sec:Generator}
	As detailed in Table~\ref{tab:Generator}, we use different solvers for different types of equations. For the diffusion-reaction equation and all types of conservation laws, we employ PDEBench \cite{takamoto2022pdebench}. The Matrix Numerical Methods (MNM) introduced in \cite{fplancksolver} are used for solving the Fokker-Planck equation, while the pseudo-spectral method from \cite{Kdvsolver} is applied to the KdV equation. For advection and wave equations, we utilize the exact solution defined by the initial conditions. The method of lines, which discretizes the PDE in space and solves the ODE in time, is used for the remaining equations.
	\begin{table}[ht]
		\small
		
		\caption{Solvers for different types of equations}
		\vskip 0.15in
		\label{tab:Generator}
		\centering
		\setlength{\tabcolsep}{1pt} 
		\renewcommand{\arraystretch}{1.5}
		\begin{tabular}{c|c}
			\hline
			Equation type & Generator   \\\hline
			Heat&\multirow{5}{*}{Method of Line}\\
			Klein-Gordon&\\
			Sine-Gordon&\\
			Porous medium &\\
			Cahn-Hilliard&\\\hline
			Diff-React&\multirow{3}{*}{PDEBench \cite{takamoto2022pdebench}}\\
			Viscous Conservation&\\
			Inviscid Conservation&\\\hline
			Advection&\multirow{2}{*}{Exact solution defined by IC}\\
			Wave&\\\hline
			\multirow{2}{*}{ Kdv}&Fourier Spectral Method\\ &\cite{Kdvsolver}\\\hline
			\multirow{2}{*}{Fokker-Planck}&Matrix Numerical Method\\ &\cite{fplancksolver}\\
			\hline
		\end{tabular}
	\end{table}

	\subsection{Symbolic Information}
	
	We illustrate the symbolic encoding of an equation in Figure \ref{fig:tree}. This encoding involves representing the equation in a tree structure, where nodes represent operations and leaves represent variables and constants. We then convert the tree structure into Polish notation \cite{Pogorzelski1965-POGRJL}. Throughout the training process, all symbols in the Polish notation are considered trainable tokens and are updated accordingly. For further details, please refer to \cite{liu2024prose,charton2022linear,dascoli2022deep,kamienny2022endtoend,tai2015improved,dyer2016recurrent}. 
	
	\begin{figure}[t]
		\centering
		\begin{tikzpicture}[scale=1]
		\tikzset{level distance=7mm}
		\tikzset{every tree node/.style={align=center,anchor=center, font=\footnotesize}}
		
		\Tree[.$+$ [.\texttt{cos} [.$\times$ {$1.5$} {$x_1$} ]]
		[.$-$ [.$\times$ $2$ $u_{x_2}$ ] {$2.6$} ]]
		\end{tikzpicture}
		\caption{Tree encoding of the example expression $\cos(1.5x_1) + 2u_{x_2} -2.6 $. }
		\label{fig:tree}
	\end{figure}
	
	\section{PROSE Model}\label{appendix:prose_details}
	
	We refer to \cite{liu2024prose,sun2025towards} for model details.
	\section{Experiment Setup}
	\subsection{Training}
	Relative squared error $\mathcal{L}^2$ is used for the data predictions. In comparison to the standard mean squared error, the relative squared error makes the learning process uniform across different types of PDE systems, as solutions of different systems may have different value ranges. \\\\
	Unless otherwise specified, the models are trained using the AdamW optimizer for 15 epochs where each epoch is 2K steps. Each task is adapted with 5 inner-loop gradient descent, and 5 tasks-specific updates before one meta update. During evaluation, each task is adapted with 20 inner-loop gradient descent. On 1 NVIDIA GeForce RTX 4090 GPUs with 24 GB memory, the pre-training takes 1 hour for regular pre-training and 6 hours for LeMON training.
	\subsection{Hyperparameters}
	
	The model hyperparameters are summarized in Table~\ref{tab:model_hyper}, and the optimizer hyperparameters are summarized in Table~\ref{tab:optim_hyper}.

	\begin{table}[!ht]
		\centering
		\small
		\setlength{\tabcolsep}{1pt} 
		\renewcommand{\arraystretch}{2}
		\begin{tabular}{l l | l l }
			\hline
			Hidden dimension for attention & 512 & Hidden dimension for FFNs & 2048\\
			Number of attention heads & 8 & Fusion attention layers& 1\\
			Data encoder attention layers & 2 & Data decoder attention layers & 8\\
			Symbol encoder attention layers & 4 & & \\
			\hline
		\end{tabular}
		\caption{\textbf{Model hyperparameters.} FFN means feedforward network.}
		
		\label{tab:model_hyper}
	\end{table}
	
	\begin{table}[ht]
		\centering
		\small
		\setlength{\tabcolsep}{2pt} 
		\renewcommand{\arraystretch}{2}
		\begin{tabular}{l l | l l }
			\hline
			Learning rate & $10^{-4}$ &Gradient norm clip & 1.0 \\
			Scheduler & Cosine &  Weight decay & $10^{-4}$\\
			Batch data size& 150 & Warmup steps & 10\% of total steps\\ 
			Batch task size& 5 & \\
			\hline
		\end{tabular}
		\caption{\textbf{Optimizer hyperparameters.}}
		
		\label{tab:optim_hyper}
	\end{table}

\end{document}